# Physics-Informed Deep Learning Model for Line-integral Diagnostics Across Fusion Devices

**Cong Wang[1,+], Weizhe Yang[2], Haiping Wang[1], Renjie Yang[1], Jing Li[1], Zhijun Wang[1], Yixiong Wei[1], Xianli Huang[3], Chenshu Hu[1], Zhaoyang Liu[1], Xinyao Yu[4], Changqing Zou[1,4], Zhifeng Zhao[1,*]**

*1 Zhejiang Lab, Hangzhou 310000, China*

*2 UNSW Sydney, Sydney, NSW, AU*

*3 ENN Science and Technology Development Co Ltd, Langfang, CN*

*4 Zhejiang University, Hangzhou, Zhejiang, CN*

**Corresponding author*

*+Contact person*

## Abstract

Rapid reconstruction of 2D plasma profiles from line-integral measurements is important in nuclear fusion. This paper introduces a physics-informed model architecture called Onion, that can enhance the performance of models and be adapted to various backbone networks. The model under Onion incorporates physical information by a multiplication process and applies the physics-informed loss function according to the principle of line integration. Prediction results demonstrate that the additional input of physical information improves the deep learning model's ability, leading to a reduction in the average relative error $E_1$ between the reconstruction profiles and the target profiles by approximately $0.84 \times 10^{-2}$ on synthetic datasets and about $0.06 \times 10^{-2}$ on experimental datasets. Furthermore, the implementation of the Softplus activation function in the final two fully connected layers improves model performance. This enhancement results in a reduction in the $E_1$ by approximately $1.06 \times 10^{-2}$ on synthetic datasets and about $0.11 \times 10^{-2}$ on experimental datasets. The incorporation of the physics-informed loss function has been shown to correct the model's predictions, bringing the back-projections closer to the actual inputs and reducing the errors associated with inversion algorithms. Besides, we have developed a synthetic data model to generate customized line-integral diagnostic datasets and have also collected soft x-ray diagnostic datasets from EAST and HL-2A. This study achieves reductions in reconstruction errors, and accelerates the development of surrogate models in fusion research.

# 1 Introduction

Deep learning has recently emerged as a pivotal tool in tokamak research, enabling the acceleration of computationally intensive tasks and the exploration of extensive experimental datasets. Since 2019, there has been a surge in research using deep learning techniques focusing on the prediction of plasma instabilities[1–5], identifying and classifying MHD instabilities and transport events in experiments[6–11] and constructing surrogate models for the first-principle simulations and transport calculations[12–17]. A novel deep learning approach was introduced for forecasting disruptions in tokamak reactors, showcasing its potential to bolster fusion energy science and the prediction of complex systems[18]. This method provides reliable, high-performance predictions across various machines by leveraging high-dimensional data and supercomputing resources. Concurrently, Physics-informed neural networks (PINNs) were proposed by M. Raissi et.al.[19] to unify deep learning with nonlinear PDE constraints to solve forward and inverse problems, bridging data-driven and physics-based modeling. PINNs have endeavored to integrate physical laws into deep learning frameworks, showcasing their potential in different areas[20–24]. In fluid dynamics, Henning Wessels et.al. introduced the Neural Particle Method (NPM), an updated Lagrangian physics-informed neural network for computational fluid dynamics[22]. Specifically, it uses the Navier-Stokes equations as constraints, ensuring that the predicted velocities and pressures satisfy these fundamental principles of fluid mechanics. In materials science, Minliang Liu et.al. proposed a generic physics-informed neural network-based constitutive model[21]. By incorporating theoretical constitutive equations into the loss function, the model ensures that the predicted stress-strain relationships remain consistent with the underlying physics of soft tissues. In thermal engineering, Darioush Jalili et.al. presented a physics-informed neural network approach for solving heat conduction and convection problems[20]. The method integrated the heat equation into the loss function, ensuring that the predicted temperature distributions complied with the conservation of energy principle.

Given the high costs and limited opportunities for plasma experiments, rapid and intuitive physical insights between experimental shots are crucial. Rapid reconstruction enables physicists and operators to make informed decisions swiftly, optimizing experimental setups and mitigating potential risks. In the pursuit of sustainable nuclear fusion energy, the rapid and accurate reconstruction of plasma profiles from line-integral measurements is essential for real-time control and decision-making in tokamak reactors, where precise knowledge of internal plasma conditions can enhance operational safety and efficiency. Current inversion methods for plasma profile reconstruction generally fall into two categories: physics-based models and data-driven surrogate models. Physics-based models, while highly accurate, incur significant computational overhead due to their reliance on detailed physical equations and simulations. This method for reconstructing these profiles is computationally intensive, often requiring complex physical models based on first principles that consider spatial and temporal dynamics, multi-physical fields, and system coupling. This complexity not only increases computational time but also limits the frequency and responsiveness of diagnostic feedback. A promising approach to address this challenge is developing data-driven surrogate models. The data-driven surrogate models, particularly machine learning, offer faster solutions[25,26]. The rapid reconstruction of the plasma profile is a pivotal challenge within the diagnostics domain, playing an indispensable role in guiding fusion physics experiments and facilitating real-time plasma control.

In recent years, considerable efforts have been directed towards utilizing deep learning-based surrogate models to achieve rapid reconstructions. Diogo R. Ferreira et al.[27] employed a deconvolutional network that receives the bolometer measurement as input (a total of 56 lines of sight from both cameras) and produces a 120×200 reconstruction of the plasma radiation profile. Additionally, they explored the potential of recurrent neural networks to predict plasma disruptions using input sequences consisting of 200 time points. Chaowei Mai et al.[26] constructed three typical neural networks, including VGG-Net, a fully affine neural network and a fully convolutional neural network, for reconstructing a two-dimensional SXR profile. The input SXR data comprise a 92 × 100 matrix and the 2D SXR tomography label image is the result of linear interpolation on a 75 × 50 (3750)-dimensional square grid, from the SXR 2D emissivity profile given by the Fourier–Bessel (F-B) code[28,29]. Zhijun Wang et al.[25] developed and trained two typical neural networks, including an up-convolutional neural network and time series neural network, to predict the reconstructions of emission profiles for the soft X-ray diagnostics of the HL-2 A tokamak. The input data, consisting of measurements taken from 40 viewing chords, is represented as a 1D vector of length 40, while the target value is a 1D vector of length 1152 derived from the SXR emissivity profiles provided by the nonstationary Gaussian process tomography (NSGPT) code. Marko Blatzheim et al.[30] demonstrated the successful reconstruction of proxies for two independent, important edge magnetic field properties given simulated heat load images on the Wendelstein 7-X divertor target plates. The input picture dimensionality is 113 × 29 and the outputs are two independent, important edge magnetic field properties. They investigated six different artificial neural network architectures, ranging from shallow and simple feed-forward fully-connected neural networks to deep Inception ResNets with 24223 to 804804 free parameters are investigated. These studies utilize classical backbone networks to construct the surrogate models; however, these models fail to integrate the physical information and principles of the diagnostic systems. F. Matos et al.[31] employed the Visual Geometry Group Net (VGG-Net)[32] and the Keras framework for deep learning. The network receives two inputs: a tomographic projection (208 SXR measurements) and a corresponding mask of ones and zeros which gives information regarding which measurements in the projection are assumed to be faulty. The network outputs are probabilities over 27 possible classes. This approach introduces additional prior knowledge into the model by informing it about potentially faulty measurements, thereby enhancing its ability to perform the classification task accurately. However, similar to other models discussed, this one also does not incorporate the underlying principles of the diagnostic system into its architecture or input data.

To summarize, these studies have made significant strides in utilizing neural networks for reconstructing and predicting plasma physical properties, yet they share several limitations. Firstly, existing models predominantly rely on classical neural network architectures, failing to embed the principles and physical information of the diagnostic systems into their model structure or input data. This omission can result in a less accurate understanding and prediction of physical phenomena. Secondly, although F. Matos et al.[31] attempt to improve classification accuracy by introducing flags for potentially faulty measurements, this approach remains superficial. It does not fundamentally address how the integration of physical knowledge could enhance both data quality and model performance. Thirdly, the accuracy of these surrogate models is contingent upon the quality of the dataset labels, ideally matching the performance of the physical models used to generate those labels.

To address these identified limitations in current methodologies, we have developed a physics-informed model architecture called Onion for line-integral diagnostic systems. This architecture fully leverages physical information and principles from diagnostic systems, thereby enhancing the capabilities of data-driven methodologies. By integrating Physical information (PI) of diagnostics through the positional encoding and incorporating physical constraints into the physics-informed loss function (PILF) that adheres to the principles of diagnostic systems, our model can incorporate the physical characteristics of diagnostic systems during the inversion process, leading to more precise and interpretable predictions. This approach circumvents the limitations of current data-driven surrogate models, which operate in a data-to-data manner without fully leveraging the underlying physical information and principles of the diagnostic system. It's important to note that for different experimental discharge scenarios, PI can include a range of knowledge like line integration, MHD and even information from other diagnostics. This study focuses exclusively on line integration-related response matrices, which represent the path length of the line of sights (LOSs) through the pixels[33]. Furthermore, this work also inherits a limitation from the previous studies that built the datasets: the employed datasets lack information on the uncertainties associated with experimental measurements.

The Onion model proposed in this work demonstrates versatility by permitting the flexible selection of advanced backbone neural network architectures as its core component. Well-known backbone network architectures used in artificial intelligence tasks include VGG[32], ResNet[34], Transformer[35], Vision Transformer[36] and so on. These backbone networks serve as the core structures responsible for extracting features from input data across various AI tasks. This flexibility ensures that our model can leverage the latest advancements in the field, thereby enhancing model performance. Moreover, we will make the training data and models openly available to the fusion community, thereby fostering the development of a model and data ecosystem within the field of nuclear fusion.

The structure of this paper is outlined as follows: **Section 2** presents the four datasets utilized in this study and evaluates the quality of the target profiles within these datasets. **Section 3** details the model architecture, including the input and output representations, the encoding of PI, the construction of the PILF, and the backbone network employed. In **Section 4**, a comparative analysis of the performance of several models is conducted, followed by a discussion. **Section 5** concludes the paper and provides an outlook on future work.

# 2 Dataset

## 2.1 Experimental datasets

In this study, we have assembled two experimental datasets from line-integral diagnostic measurements obtained from two tokamak facilities: EAST and HL-2A, named Exp_EAST and Exp_HL-2A.

The Exp_EAST is compiled by Chaowei Mai[26]. **Figure 1**(a) provides an overview of the basic information regarding the SXR diagnostics. The cameras are equipped with 46 detectors each for the Upper (U) and Lower (D) arrays, and 30 detectors for the Vertical (V) array, all capable of delivering independent SXR measurements. The raw data extracted from the EAST SXR cameras, specifically U and D, serve as inputs for the dataset, which comprises 92 diagnostic data points. The corresponding target profiles consisting of 75×50 grids are images generated from 2D SXR Tomography (SXT), which are produced using F-B codes[28,29].

The Exp_HL-2A dataset was constructed by Zhijun Wang[25]. The experimental configuration of the SXR diagnostic for HL-2A is depicted in **Figure 1**(b). A total of 40 LOSs from Cameras No. 3 and No. 4 were utilized as inputs; each camera is equipped with 20 Si-PIN photon-diode detectors. The target profile, consisting of 36×32 grids, represents a 2D profile image that is derived from the SXR emissivity profiles calculated by the NSGPT code[37].

The Exp_EAST dataset encompasses both steady discharge and MHD-unstable discharge data, whereas the Exp_HL-2A dataset is limited to steady discharge data. Regrettably, neither dataset includes uncertainty information for the measured signals or the SXR emissivity profiles. This omission precludes the conduct of uncertainty-related research and analysis within the scope of this work.

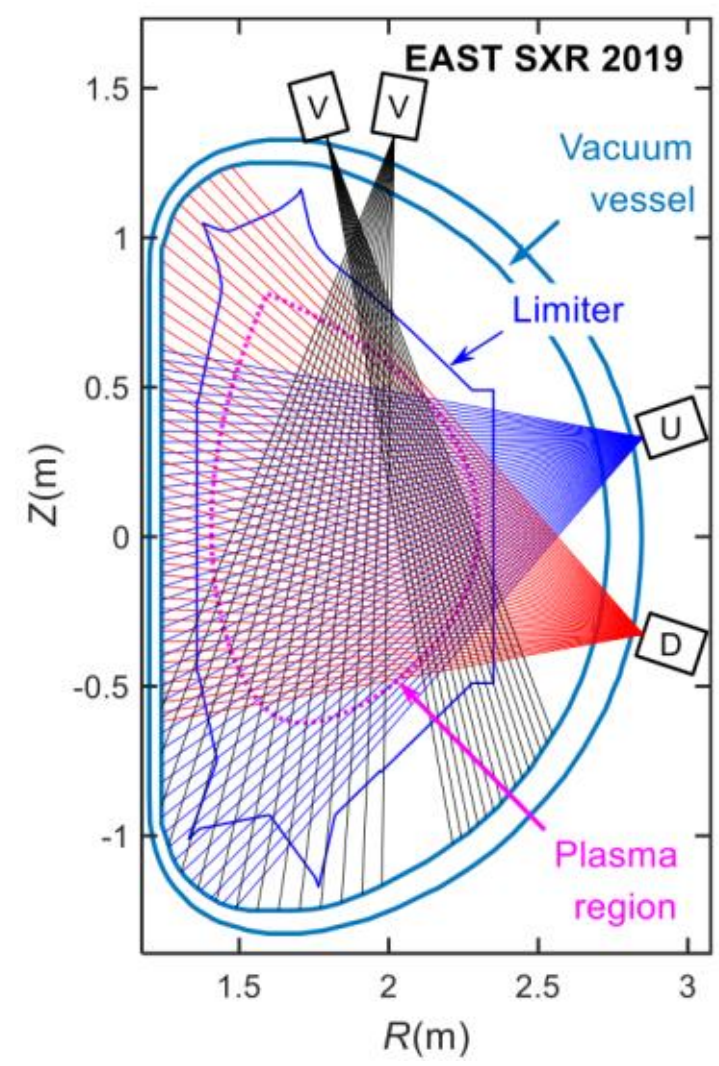

(a)

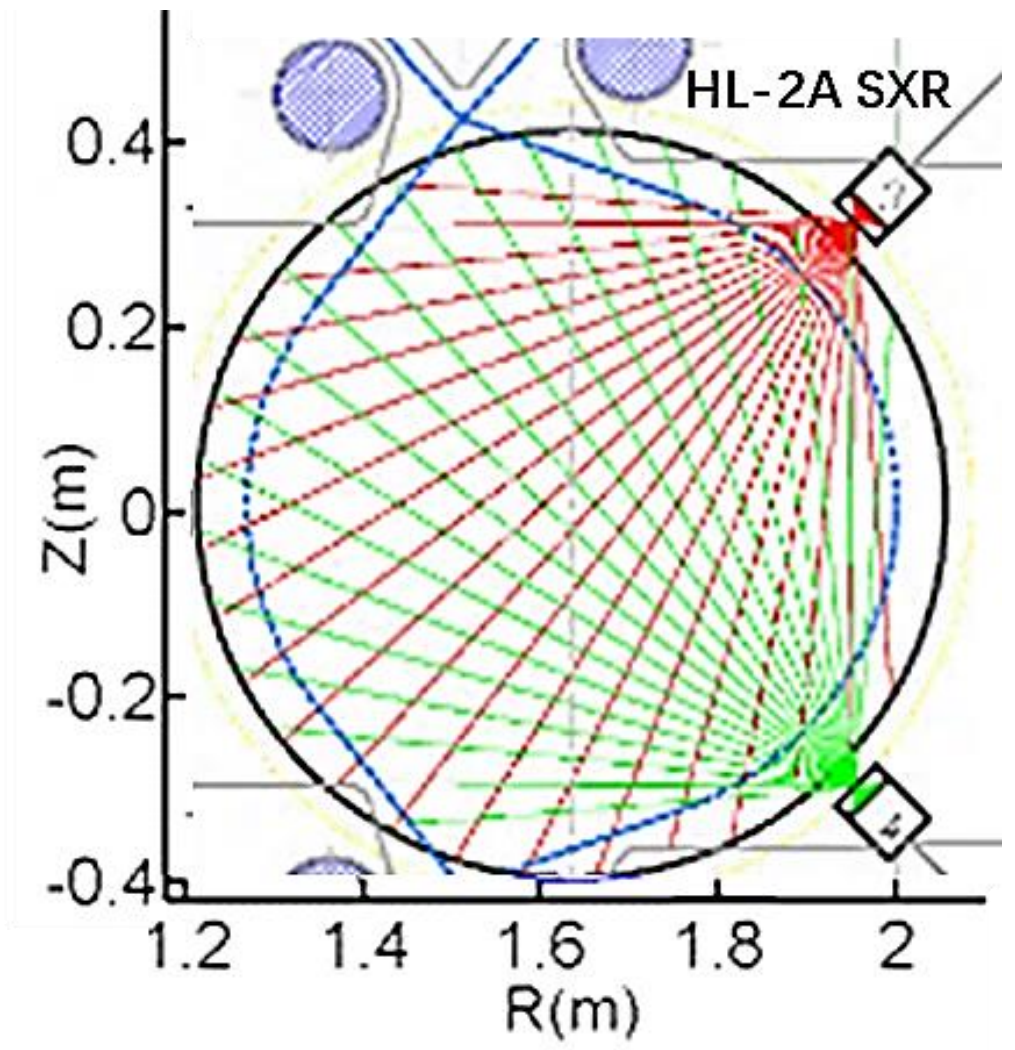

(b)

**Figure 1**: (a) EAST cross section and three cameras (U, D and V) of the SXR diagnostic with locations of their LOSs shown (three detector sightlines); (b) The experimental configuration of the SXR diagnostic for HL-2A.

## 2.2 Synthetic datasets

Additionally, we have developed a synthetic data model, designed for the generation of synthetic data including target profiles and virtual diagnostic data for line-integral diagnostics. This model creates simple 2D profiles (target profiles) of arbitrary region sizes and leverages a forward line-integral model to create their corresponding virtual line-integral diagnostic data. The model allows for parameter customization based on the characteristics of the actual line-integral diagnostic for different tokamak devices. The flowchart of the synthetic data model is illustrated in the **Figure 2**. Parameters such as the number of radial grid points ($r_{num}$), the number of vertical grid points ($z_{num}$), and the response matrix ($R$) can be set accordingly. The model randomly selects a central point within the grid area and initializes it with a maximum value. Subsequently, values are assigned to each grid point according to its positional relationship with the central point and following the assignment rule that the value decreases uniformly with increasing distance from the central point, thus obtaining a complete 2D profile. By applying the principle of line integration outlined below, the model generates the virtual diagnostic data.

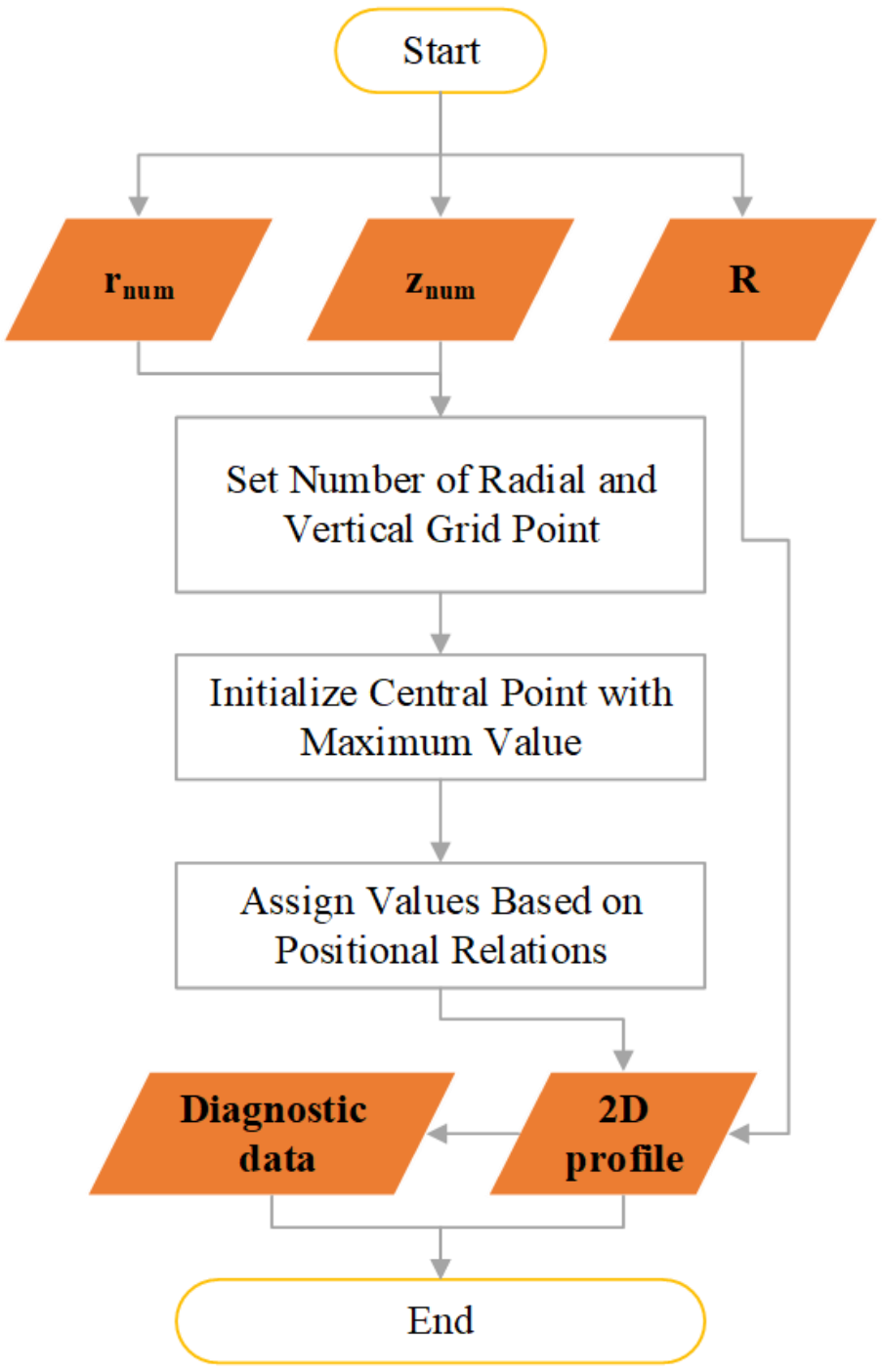

**Figure 2**: The flowchart of the synthetic data model. The inputs to the model include the number of radial grid points ($r_{num}$), the number of vertical grid points ($z_{num}$), and the response matrix ($R$). The outputs of the model include the diagnostic data and 2D target profiles.

The principle of the line-integral diagnostic is encapsulated in Equation (2-1)

$$x_i = R_i \cdot y + \Delta_i \tag{2-1}$$

, where $x_i$ represents the diagnostic data of the $i$-th chord. The response matrix for the i-th chord is denoted by $R_i$, and $y$ is the target profile. $R_i \cdot y$, named back-projection[31] (BP), is the projection of the 2D profile back into the detector's measurement. It calculated by considering the starting and ending positions, as well as the beam width of the LOS. The term $\Delta_i$ accounts for the systematic and statistical errors, while for synthetic data, $\Delta_i$ is set to 0. Compared to experimental measurements and 2D profiles obtained by inversion algorithm, the synthetic data model produces data that is strictly free from algorithm errors including systematic error and random error.

The synthetic datasets of the SXR diagnostic for EAST and HL-2A, named Synthetic_EAST and Synthetic_HL-2A, are generated using the synthetic data model. The examples of the generated target profiles are presented in **Figure 3**. It is worth noting that the 2D profile is a simple circular profile, which is fundamentally different from the profiles obtained by inversing experimental data. The synthetic data are merely intended to construct a dataset that free from the inherent errors comparing with experimental datasets.

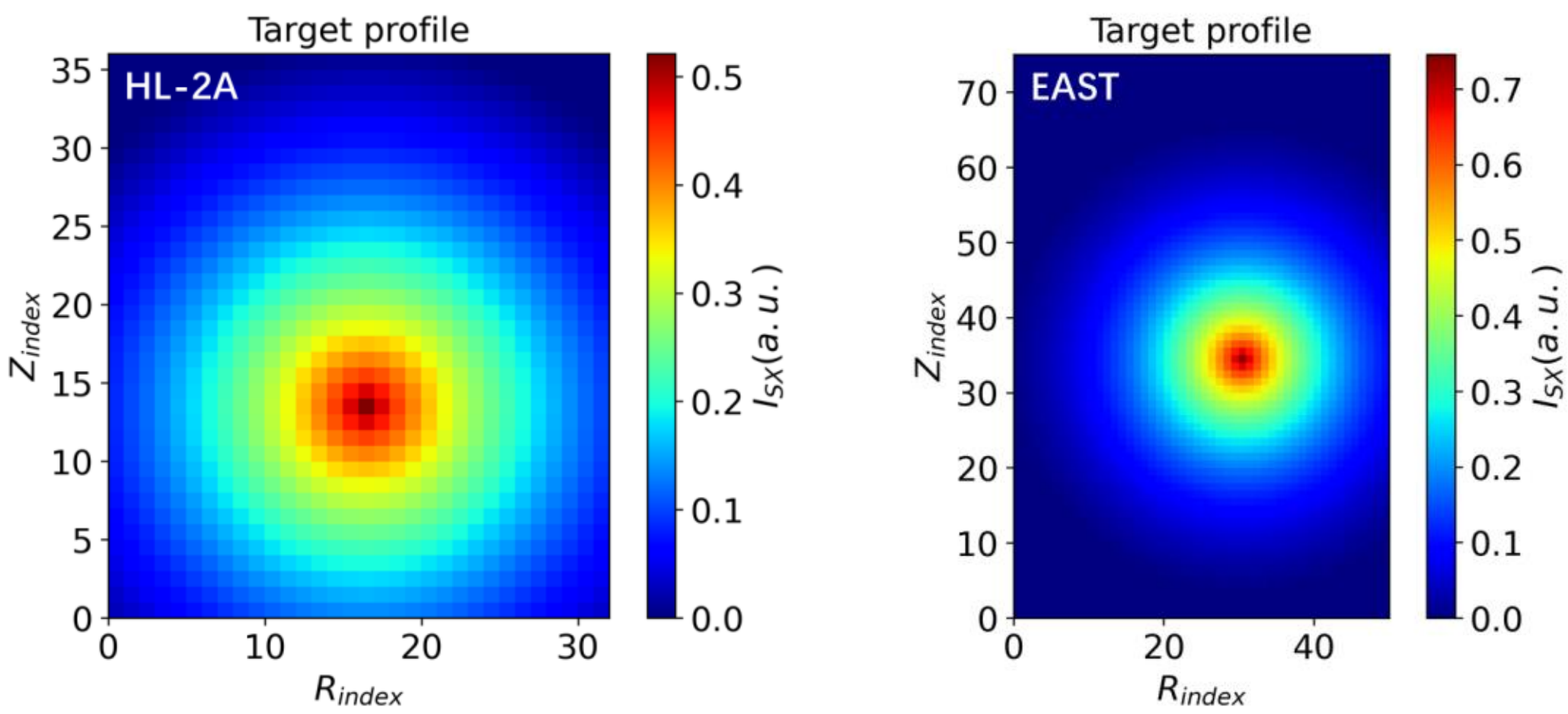

(a) (b)

**Figure 3**: Examples of the synthetic target profiles. (a) Example of target profile for Synthetic_HL-2A. (b) Example of target profile for Synthetic_EAST.

## 2.3 Evaluation of datasets

Based on the principle of line integration, the quality of the k-th sample can be assessed using mean relative error $\varepsilon_k$ of BPs, as shown in Equation (2-2).

$$\varepsilon_k = \frac{1}{n}\sum_{i=1}^{n}\left|\frac{x_i^k - R_i \cdot y^k}{x_{max}^k}\right| \qquad 2\text{-}2$$

$x_{max}^k$ is the maximum value of diagnostic data $x^k$ and $n$ is the number of LOSs. Hence, the inherent error of the dataset can be assessed by $\bar{\varepsilon}$, as shown in Equation (2-3)

$$\bar{\varepsilon} = \frac{1}{m}\sum_{k=1}^{m}(\varepsilon_k) \qquad 2\text{-}3$$

, where $m$ is the number of samples in the dataset. $\bar{\varepsilon}$ also indicates the goodness of various inversion algorithms. **Figure 4** presents the comparison between BPs and diagnostic data of k-th sample across different datasets. For the samples from the experimental datasets shown in **Figure 4**(a) and **Figure 4**(b), there is a deviation between the diagnostic data (blue triangles) and the BPs obtained based on the target profile (red dots), which means that the target profile generated by the inversion algorithm in the dataset itself has errors, and this error comes from the inversion algorithm. For the samples from the synthetic datasets shown in **Figure 4**(c) and **Figure 4**(d), the diagnostic data and the BPs are in good agreement, verifying that the synthetic data model produces data that is strictly free from algorithm errors.

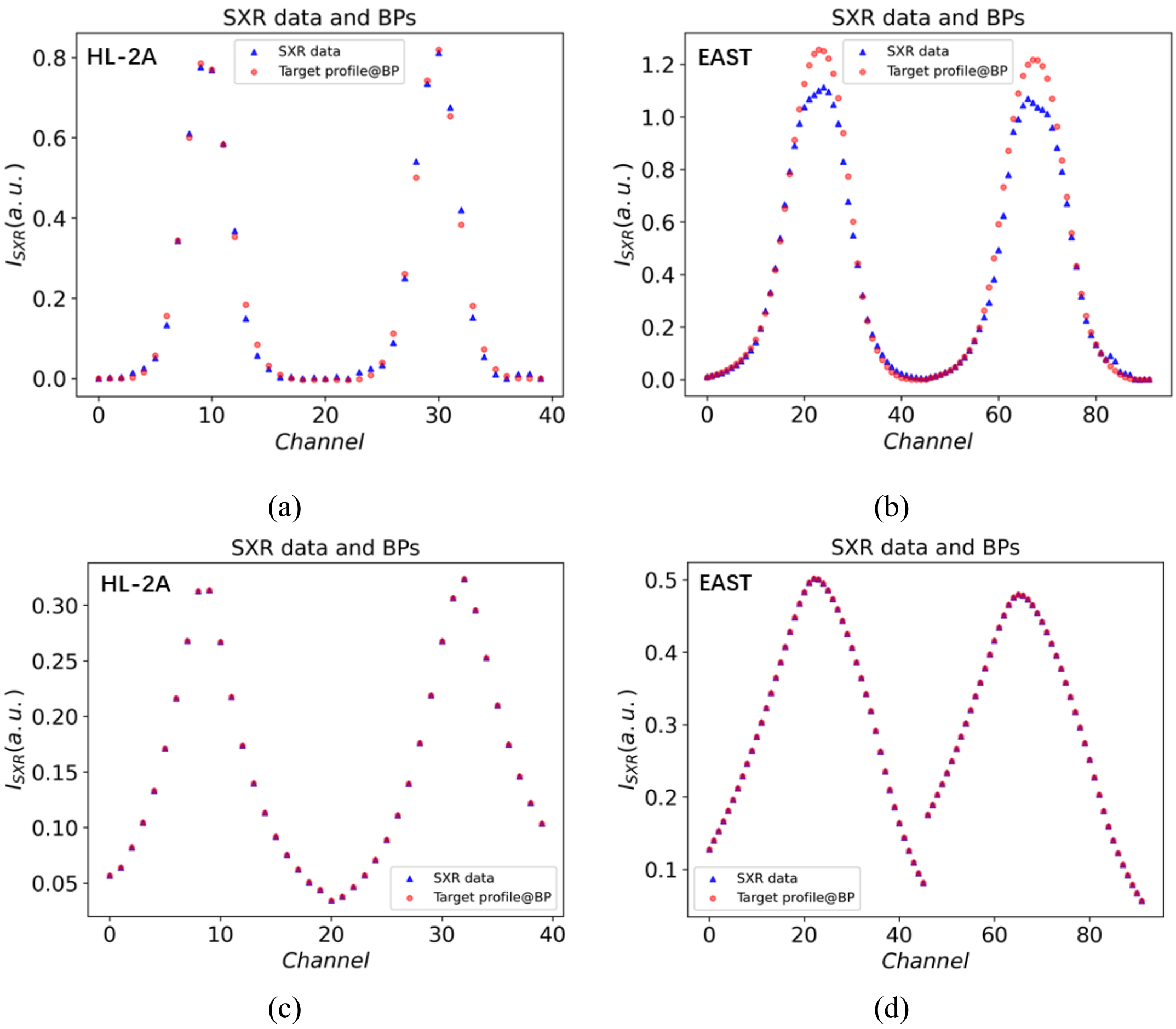

**Figure 4** Comparison between BPs and diagnostic data of k-th sample across different datasets, (a) for Exp_HL-2A, (b) for Exp_EAST, (c) for Synthetic_HL-2A, and (d) for Synthetic_EAST. The blue triangles are the SXR data of LOSs, and the red circles are the BPs calculated based on the target profile.

**Table 1** presents an overview of the basic information for four datasets. For datasets of EAST, the input is the diagnostic data of 92 LOSs, and the target profile is a 75*50 matrix. For datasets of HL-2A, the input is the diagnostic data of 40 LOSs, and the target profile is a 36*32 matrix. The number of samples in the Synthetic_EAST, Synthetic_HL-2A, Exp_EAST and Exp_HL-2A are 100000, 100000, 43378, and 765200 respectively. This work split the dataset into training, validation, and test sets in a ratio of 70%, 20%, and 10%, respectively. The $\bar{\varepsilon}$ of Synthetic_EAST and Synthetic_HL-2A are close to 0, which means the data rigorously adheres to the principles of line integration. The $\bar{\varepsilon}$ of Exp_EAST and Exp_HL-2A are $4.96 \times 10^{-2}$ and $5.42 \times 10^{-2}$, respectively.

**Table 1** An overview of the basic information for datasets including input size, profile size, number of training set, number of validation set, number of test set and $\bar{\varepsilon}$ of datasets.

| Dataset | Input size | Profile size | Train_num | Valid_num | Test_num | Error ($\bar{\varepsilon}$) |
|---|---|---|---|---|---|---|
| Synthetic_EAST | 92 | 75*50 | 70000 | 19999 | 10001 | $3.70 \times 10^{-8}$ |
| Synthetic_HL-2A | 40 | 36*32 | 70000 | 19999 | 10001 | $2.54 \times 10^{-8}$ |
| Exp_EAST | 92 | 75*50 | 30364 | 8676 | 4338 | $4.96 \times 10^{-2}$ |
| Exp_HL-2A | 40 | 36*32 | 535640 | 153040 | 76520 | $5.42 \times 10^{-2}$ |

# 3 Model Architecture

The goal of this project is to enhance the performance of the backbone neural network by developing a physics-informed model architecture, which aims to improve the predictive capability of surrogate models for line-integral diagnostic systems. The model utilizes measurement signals from the diagnostic system as input and the outcomes of inversion algorithms as target profiles. The training process incorporates four datasets: two are synthetically generated using the synthetic data model (Synthetic_EAST and Synthetic_HL-2A), while the other two are obtained from experimental data (Exp_EAST and Exp_HL-2A).

**Model Design:** The physics-informed model is meticulously designed with several key components shown in **Figure 5**: an input representation layer to transfer Input information to higher dimensions for better feature extraction and easy merging with PI; a PI feature extraction side chain to extract diagnostic PI, and feed it into the following neural network; a backbone network layer that offers the flexibility to select a high-performance neural network structure as the core; an output representation layer, which consists of two fully connected (FC) layers for final output computation; and the PILF, which is a combination of mean square error, a physical constraint loss derived from the principles of line integration, and an L2 regularization term. Each component of the loss function will be described in detail in **Section 3.3**, including its explicit mathematical formulation. Each component of the model will be described in detail in the subsections of **Section 3**.

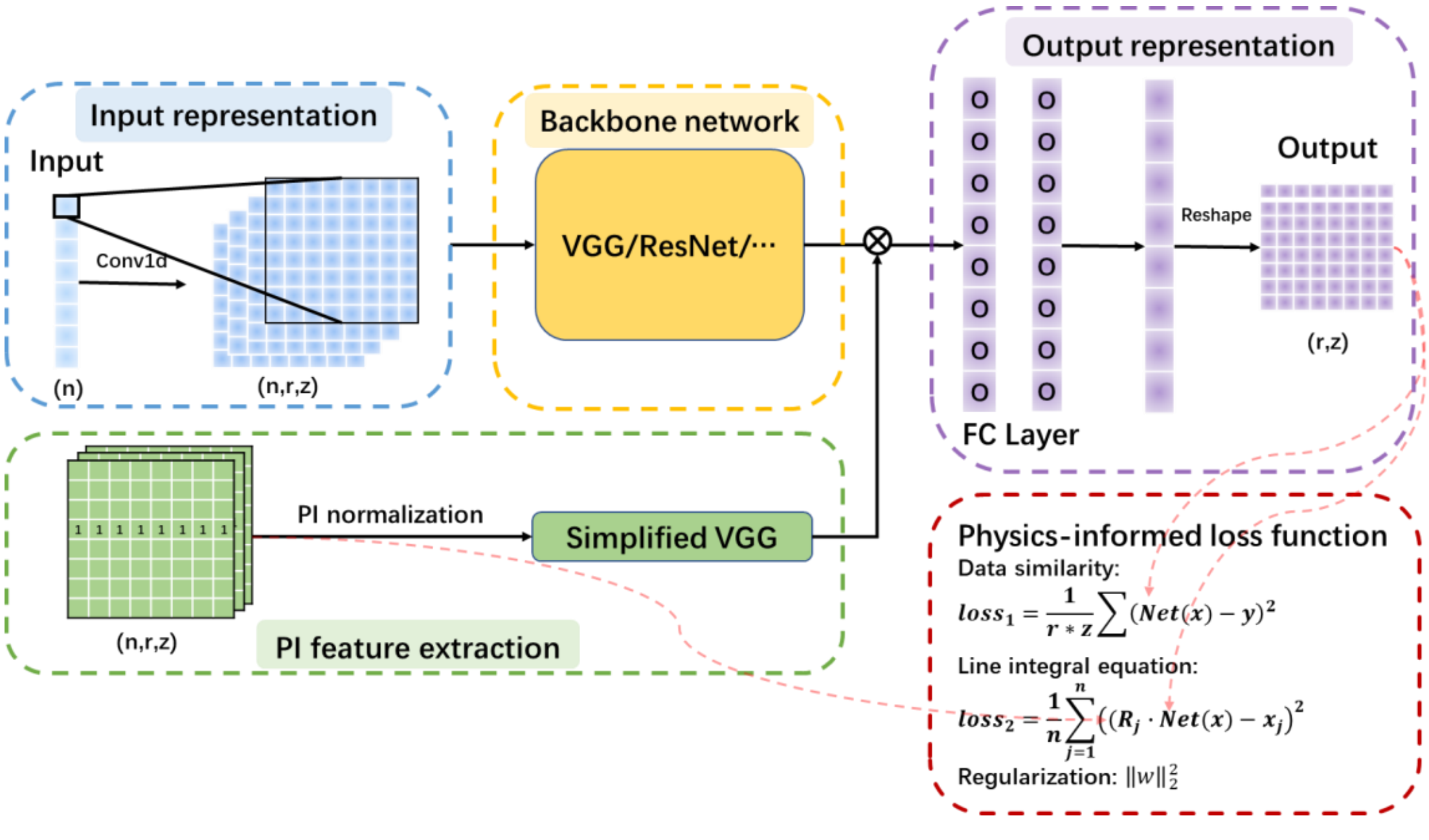

**Figure 5** Physics-informed model including input representation (blue dotted frame), backbone network (yellow dotted frame), output representation (purple dotted frame), PI feature extraction (green dotted frame) and PILF (red dotted frame).

**Training Details:** We employ the Adam optimizer with an initial learning rate of 0.0001. To further refine the training process, we have implemented a cosine annealing learning rate scheduler to adjust the learning rate. This strategy operates over a period of 50 epochs, during which the learning rate is gradually reduced to a minimum value of 0.00001. The cosine annealing method[38–40] is particularly effective in managing the convergence rate, as it helps to avoid potential stalls in training progress that can occur with static or prematurely reduced learning rates. The role of the cosine annealing learning rate scheduler and its interaction with the Adam optimizer are detailed in Appendix A. To ensure a fair comparison of performance across all models by controlling the number of training epochs, the training process is executed on an NVIDIA V100 GPU and encompasses a total of 50 epochs. The hyperparameters are set with a batch size of 256 and a weight decay (L2 regularization) coefficient of 0.0001. The loss curves for each model are detailed in Appendix B.

## 3.1 Input and output representation

The input representation transforms the diagnostic data, originally of size $(n)$, into a three-dimensional format $(n, z, r)$. $z$ and $r$ are the numbers of grids in the Z and R directions of profiles. For different datasets, $n, z, r$ take different values; for Synthetic_EAST and Exp_EAST, they are 92,75,50, and for Synthetic_HL-2A and Exp_HL-2A, they are 40,36,32. Prior to being fed into the backbone network, the input data passes through a 1D convolutional layer that increases the number of each input channel from 1 to $r \times z$. The resulting tensor is then reshaped into a format of $(n, z, r)$ for further processing. This design serves dual purposes: it aligns with the input shape requirements of the main network and enhances the feature representation capabilities of the input data. This conversion enables the model to capture the local correlations and spatial structures present in the input data more effectively.

The output from the backbone network is initially flattened to integrate all local features into a single global feature vector. Subsequently, this vector is processed through two FC layers, which facilitates the learning of global relationships among the features, and the output length is $r \times z$. The input dimension of the first FC layer is the length of the vector, and its output dimension is of size $r \times z$. The second FC layer preserves the dimensionality, meaning the input and output dimensions are identical. For the final two FC layers, either ReLU or Softplus activation functions are employed. For other hidden layers, the ReLU activation function is uniformly used to introduce nonlinearity. The comparison between the two activation functions is depicted in **Figure 6**. The ReLU activation function is defined as $max(0, x)$. For $x < 0$, the derivative of the ReLU activation function drops to zero abruptly. The Softplus activation function is considered a smoothed version of the ReLU activation function. For $x < 0$, the derivative of the Softplus activation function gradually decreases, approaching zero.

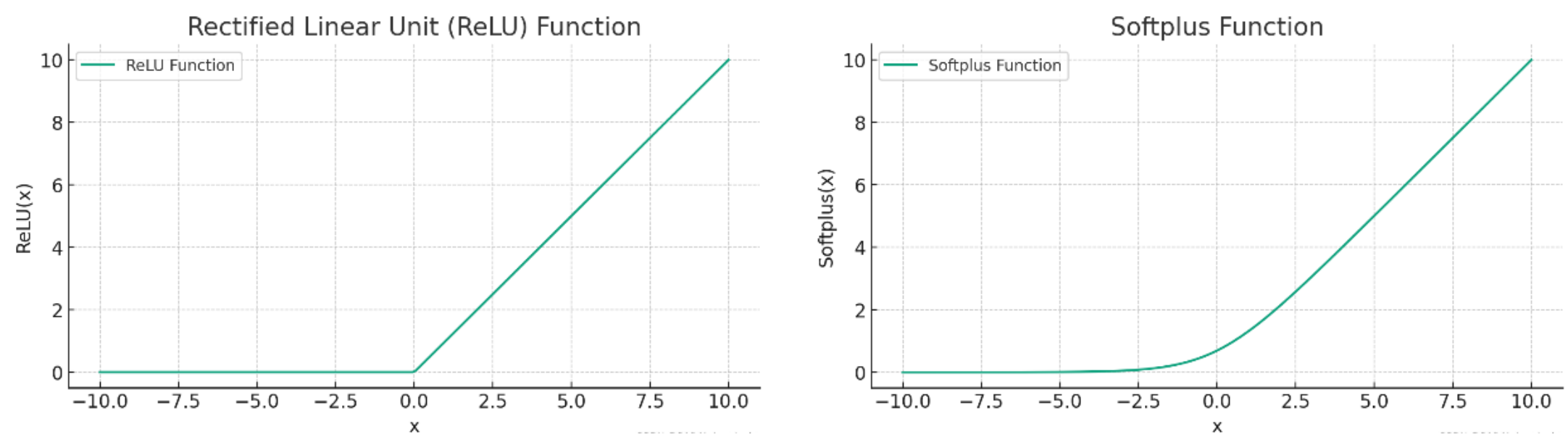

(a) (b)

**Figure 6**: Comparison between the two activation functions: (a) ReLU activation function; (b) Softplus activation function.

## 3.2 Physical information feature extraction

PI feature extraction (green dotted frame) is depicted in **Figure 5**. The inspiration for introducing PI comes from the work of Tailin Wu[41] which introduces the boundary state in the model. The PI of the line-integral diagnostic system primarily consists of LOS information which refers to the response matrix of LOSs within the diagnostic system. Each LOS's response matrix is represented as a matrix of size $(z, r)$, consistent with the dimensions of the target profile. Therefore, for a device with $n$ LOSs, the size of PI is $(n, z, r)$. **Figure 7** displays the grayscale images of the response matrices for several LOSs of the EAST device, as well as the grayscale image obtained by summing the response matrices of all 92 LOSs. **Figure 8** presents the grayscale images of the response matrices for several LOSs of the HL-2A device, the grayscale image derived from the sum of the 40 LOSs' response matrices. The lines in the 2D plots represent the path trajectories of the LOSs, while the brightness indicates the path length traversed through the pixels.

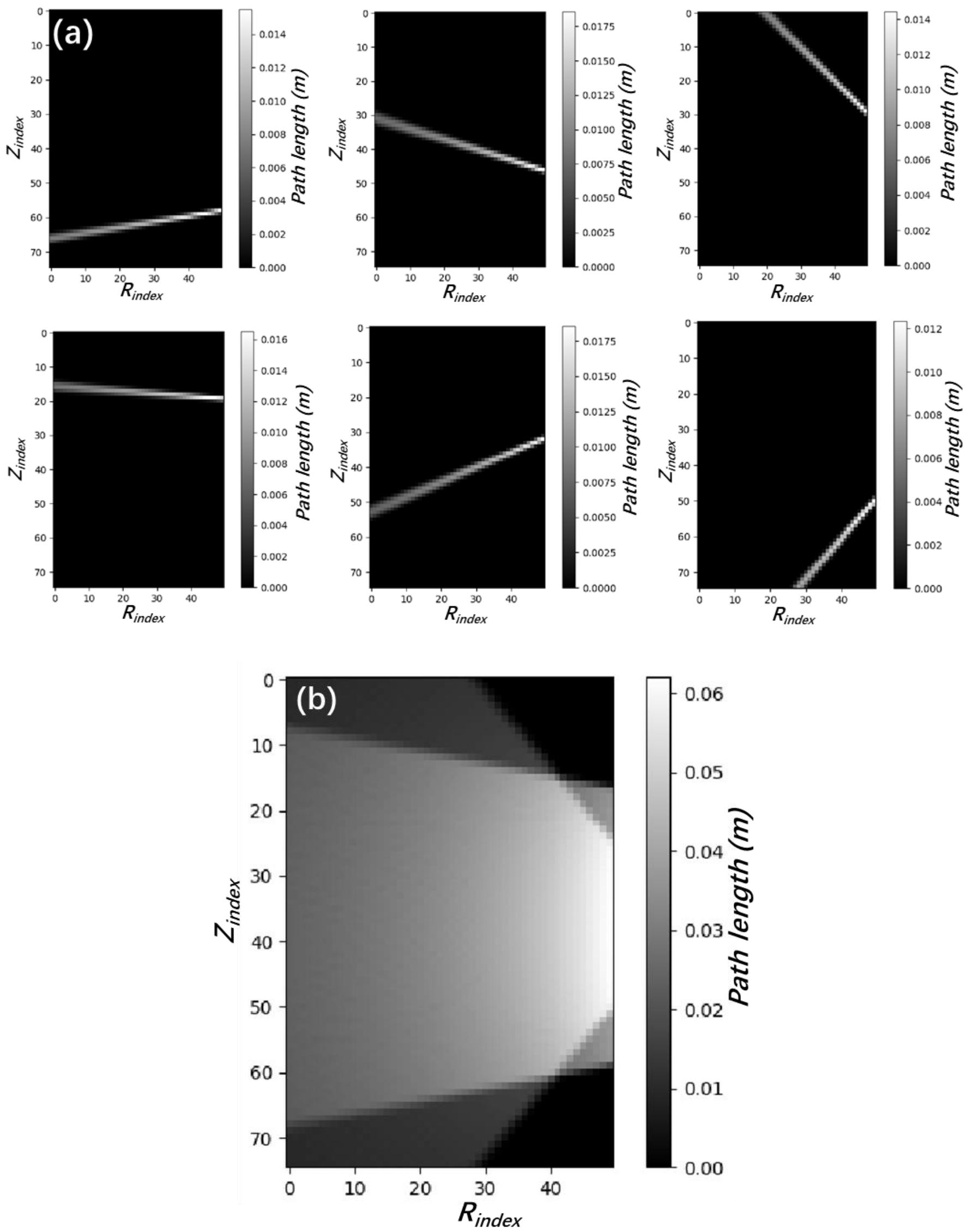

**Figure 7**: (a) The grayscale images of response matrices for several channels of the EAST device; (b) The grayscale image obtained by summing the response matrices of all 92 channels.

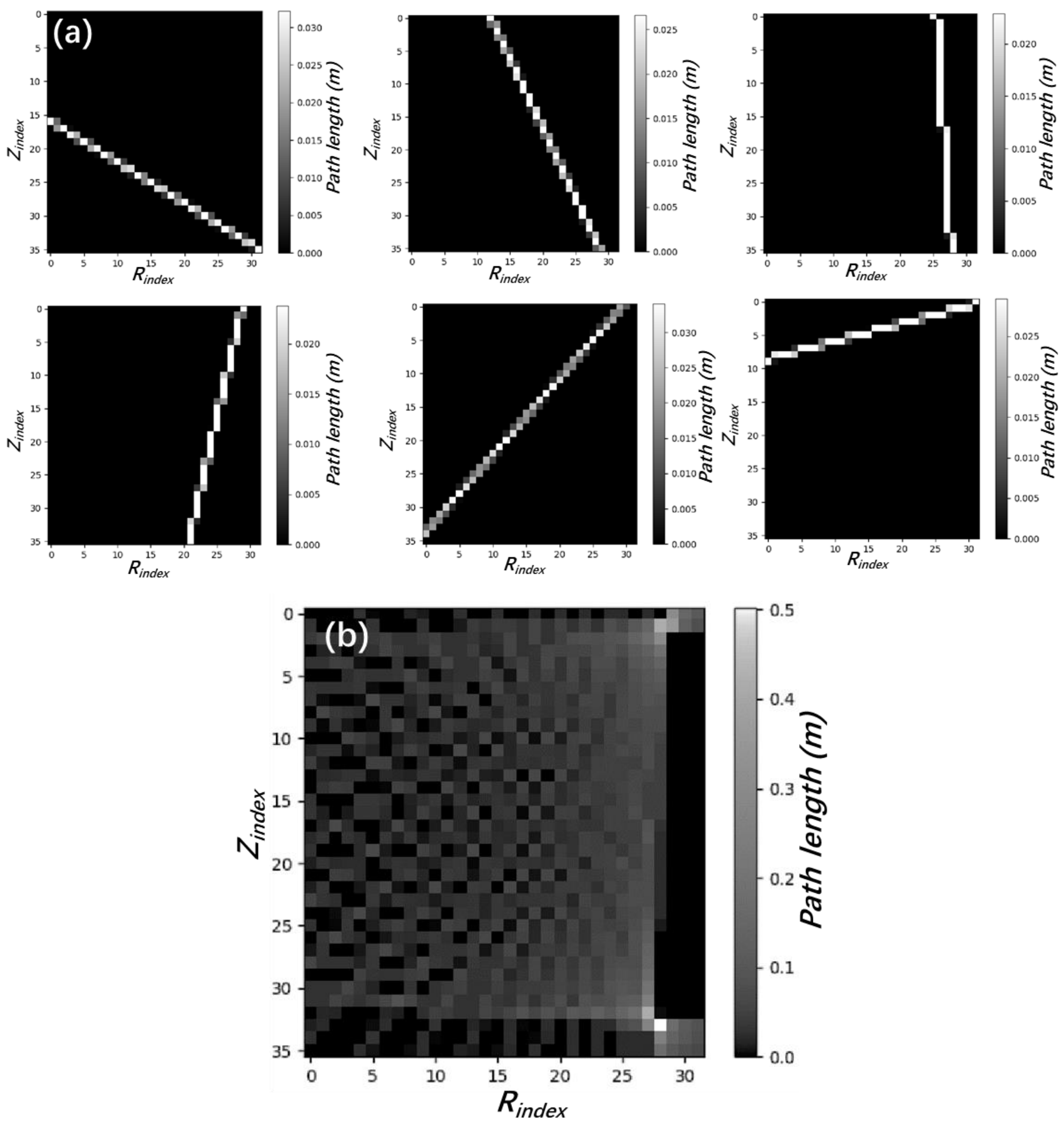

**Figure 8**: (a) The grayscale images of response matrices for several channels of the HL-2A device; (b) The grayscale image obtained by summing the response matrices of all 40 channels.

Then, the PI undergoes positional encoding, which includes normalization, prior to being processed by the simplified VGG-Net present in **Figure 9**. For the standard VGG-Net[32], it has at least 11 hidden layers, whereas the simplified version has only 6 layers. The input data passes through Conv_block1 to produce $2n$ output channels, resulting in an output tensor of shape $(2n, z, r)$. Subsequently, a MaxPool2d layer with a $2 \times 2$ kernel downsamples the spatial dimensions, reducing the feature map size while retaining the most salient features. The resulting tensor has a shape of approximately $(2n, z/2, r/2)$. Next, Conv_block2 further processes the data to generate $4n$ output channels, extracting more complex features. The output tensor now has a shape of $(4n, z/2, r/2)$. Another MaxPool2d layer is then applied to further downsample the spatial dimensions, resulting in a tensor shape of $(4n, z/4, r/4)$. Finally, Conv_block3 captures even higher-level features producing $8n$ output channels. The output tensor has a shape of $(8n, z/4, r/4)$. An AdaptiveMaxPool2d layer adjusts the spatial dimensions to a fixed size of $3 \times 3$, ensuring consistent feature map dimensions regardless of the input size. The final output tensor shape is $(8n, 3,3)$. The detail introduction of expansion unit, standard unit, MaxPool2d and AdaptiveMaxPool2d are present in **Section 3.4**.

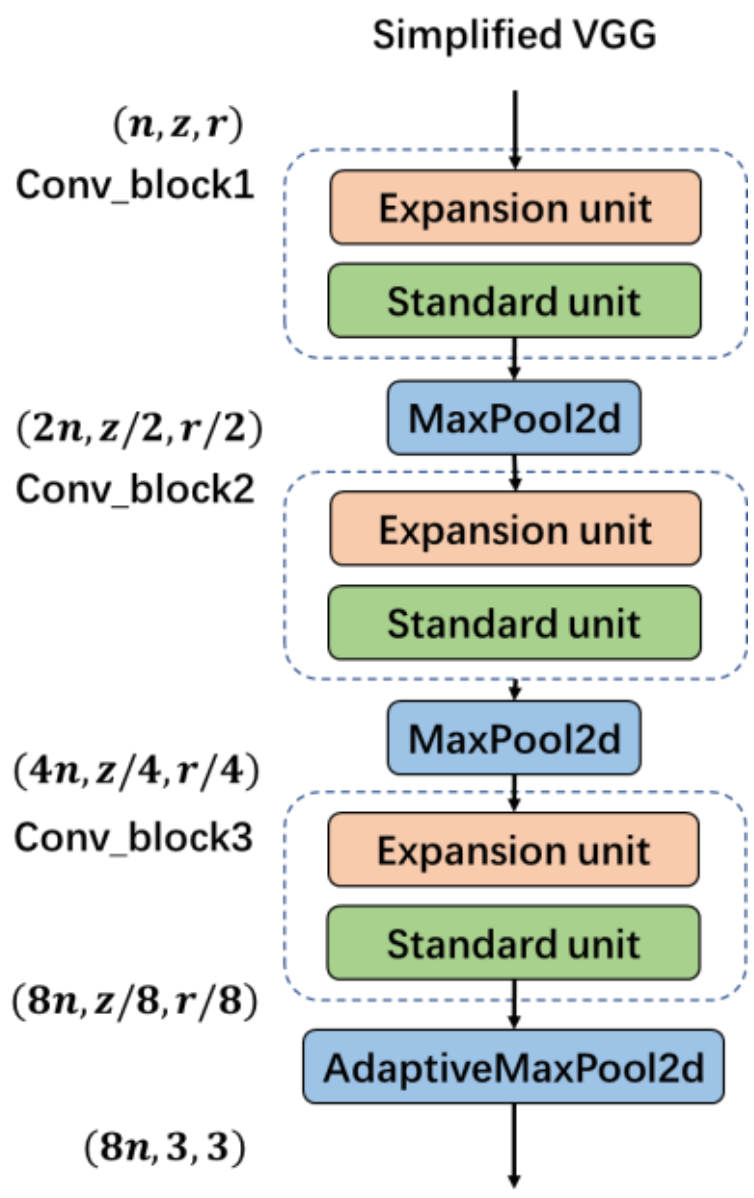

**Figure 9** The simplified VGG-Net including the expansion unit, standard unit, MaxPool2d and AdaptiveMaxPool2d.

The features extracted in the side chain are then integrated with the output of backbone network through the element-wise multiplication, facilitating the fusion of information in a manner that enhances the model's predictive capabilities. We provide a simple example to elucidate the rationale behind designing the element-wise multiplication process that enables the model to capture the complex dependencies between PI and the inputs. For a neural network node $f(\cdot)$, $x$ represents input from varying samples and changes dynamically, whereas $x'$ is the PI which is a constant for the same device. When $x$ and $x'$ are combined via addition or concatenation into the model, the outputs are given by Equation (3-1) and Equation (3-2).

$$f(x + x') = \omega_1(x + x') + b_1 \quad \text{3-1}$$

$$f([x, x']) = \omega_2 x + \omega_2' x' + b_2 \quad \text{3-2}$$

$\omega_1$ and $b_1$ are the weight and bias for $f(x + x')$. $\omega_2$, $\omega_2'$ and $b_2$ are the weight and bias for $f([x, x'])$. The $x$ and $x'$ are involved in computations either through addition (Equation (3-1)) or as independent variables (Equation (3-2)), that allows them to be considered as separate influencing factors, i.e., $\omega_1 x$ and $\omega_1 x'$ in Equation (3-1), and $\omega_2 x$ and $\omega_2' x'$ in Equation (3-2). However, empirical testing of both approaches revealed suboptimal model performance.

When $x$ and $x'$ are combined using element-wise multiplication and input into the model, the resulting output is described by Equation (3-3).

$$f(x \times x') = \omega(x \times x') + b \quad \text{3-3}$$

$\omega$ and $b$ are the weight and bias for $f(x \times x')$. By multiplying $x$ and $x'$ , their interaction effect is directly modeled within the function, potentially capturing more complex dependencies between the $x$ and $x'$. This method of combination can better account for the nuanced interactions between $x$ and $x'$, leading to a richer representation of their relationship within the neural network.

## 3.3 Physics-informed loss function

The PILF is composed of three distinct components shown in Equation (3-4) to Equation (3-6), each serving a specific purpose in the model's training process.

$$loss_1 = \frac{1}{r*z}\sum(Net(x) - y)^2 \quad 3\text{-}4$$

$$loss_2 = \frac{1}{n}\sum_{j=1}^{n}\left((R_j \cdot Net(x) - x_j\right)^2 \quad 3\text{-}5$$

$$L2 = \|w\|_2^2 \quad 3\text{-}6$$

$Net(x)$ is the reconstruction profile with the input $x$ and the physics-informed model $Net(\cdot)$. $y$ is the target profile and $w$ is the weight vector. Firstly, the discrepancy between the $Net(x)$ and $y$ quantifies the accuracy of the model's predictions in $loss_1$. Secondly, $loss_2$ includes the error between the BP obtained by the response matrix of $j$-th LOS $R_j$ times $Net(x)$ and the actual input of $j$-th LOS $x_j$, ensuring that the model's internal physics are consistent with the inputs. $n$ is the number of LOSs. Lastly, an L2 regularization term is introduced to prevent overfitting by penalizing excessive complexity in the model's parameters. The weighting coefficient of each component is collectively defined in Equation (3-7).

$$L = w_1 \cdot loss_1 + w_2 \cdot loss_2 + \lambda \cdot L2 \quad 3\text{-}7$$

$$w_1 = 1.0, w_2 = c_1 \frac{loss_1}{loss_2} \quad 3\text{-}8$$

We introduce simple weighting coefficients, denoted as $w_1$ and $w_2$ for the first two components in Equation (3-7). The $w_1$ is equal to 1.0 and the $w_2$ is the ratio of $loss_1$ to $loss_2$ multiplied by the coefficient $c_1$, where the ratio of $loss_1$ to $loss_2$ is to eliminate the magnitude difference of the two losses and $c_1$ represents the relative attention paid by the model to $loss_2$. When $c_1$ equal to 1.0, it means the model treats both losses as equally important in each epoch of training. This design enables the loss function to dynamically adjust the weighting contributions of $loss_2$ based on their relative magnitudes at any given time. The coefficient $\lambda$ for the L2 regularization term is set to 0.0001. By doing so, the model training process can adaptively emphasize the components that require more refinement, ensuring a more effective optimization trajectory.

## 3.4 Backbone network

The choice of backbone network is arbitrary and can range from established architectures such as VGG[32], ResNet[34], Transformer[35], Vision Transformer[36]. In this work, we have constructed two backbone networks shown in **Figure 10**, VggOnion and ResOnion, based on VGG and ResNet architectures, respectively.

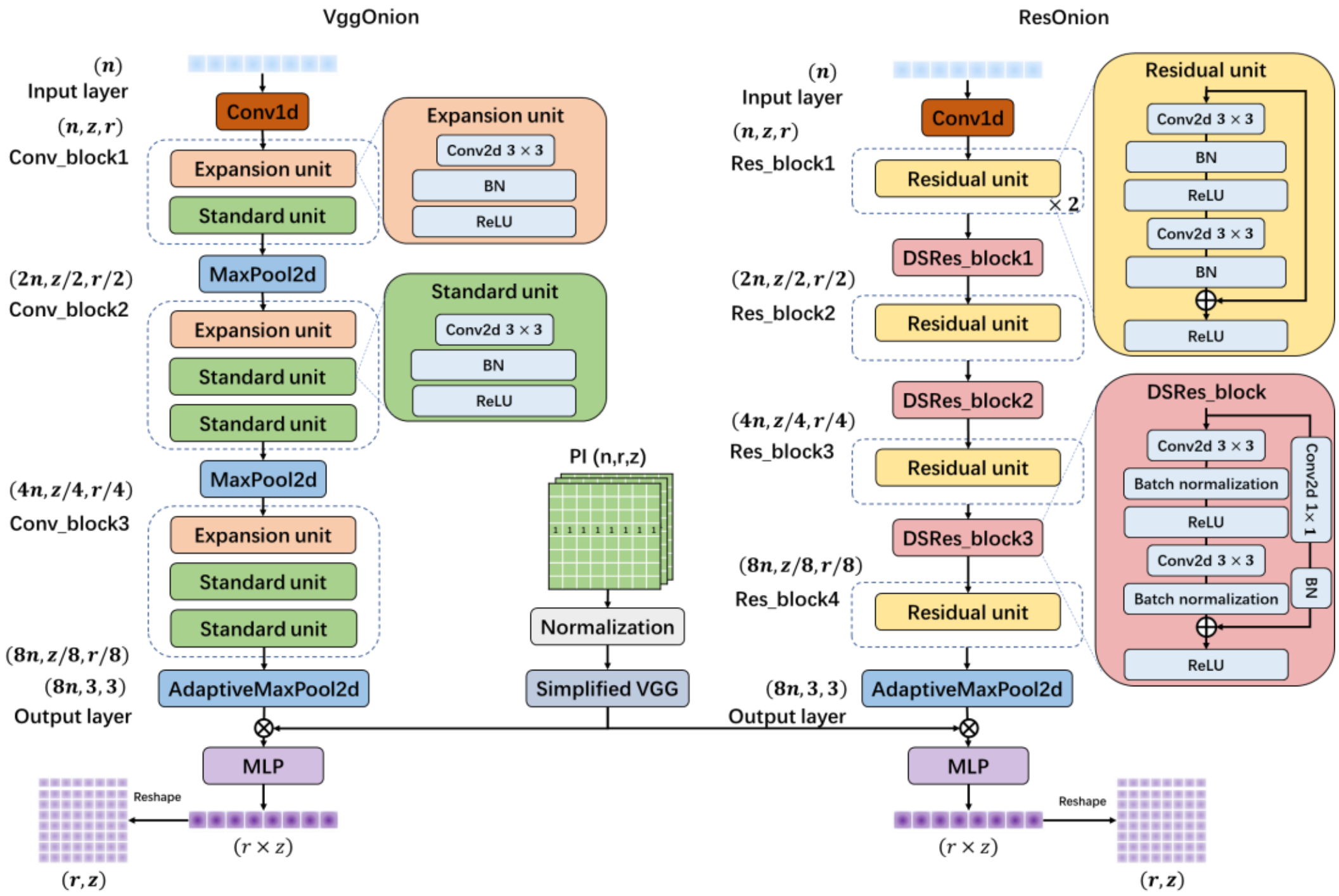

**Figure 10** Architectures of VggOnion and ResOnion.

The VggOnion incorporates 8 convolutional layers, consisting of 3 convolutional blocks (Conv_blocks). The Conv_block architecture consists of a fixed number of convolutional units designed to progressively extract and refine spatial features from the input data. The block begins with a convolutional expansion unit followed by one or more standard convolutional units. For the convolutional expansion unit, it initiates with a convolutional layer that effectively doubles the depth of the feature representation. This layer employs a kernel size of $3 \times 3$ with padding of $1$ on each side. A batch normalization (BN) layer follows, maintaining the stability of the training process by normalizing the convolved features across mini-batches. A ReLU activation layer is then applied to introduce non-linearity into the model. For the standard convolutional unit, it includes a convolutional layer keeping the depth of the feature representation with a kernel size of $3 \times 3$ with padding of $1$, a batch normalization layer and a ReLU activation layer. Between the Conv_blocks, there is a max pooling layer characterized by a kernel size of $2 \times 2$ and a stride of $2$. This operation reduces the spatial dimensions of the feature maps by half, thereby decreasing the computational load for subsequent layers while retaining the most salient features through the selection of maximum values within each pooling window. Finally, an AdaptiveMaxPool2d layer adjusts the spatial dimensions to a fixed size of $3 \times 3$.

The ResOnion network incorporates 16 hidden layers, consisting of 4 residual blocks (Res_blocks), and 3 downsampling residual blocks (DSRes_blocks). The Res_block is designed based on the principles of residual learning, which addresses the problem of vanishing gradients in deep neural networks by facilitating the training of deeper architectures. The Res_block consists of residual units containing two convolutional layers using a kernel size of $3 \times 3$ and padding of $1$, each followed by a batch normalization layer and a ReLU activation layer. Before the last ReLU activation layer, the transformed input and the original input are summed. The DSRes_block incorporates both residual learning and dimensionality scaling mechanisms to effectively handle feature maps with varying channel depths and spatial resolutions. It includes a side chain used to match the dimensions of the original input to transformed input. It employs a $1 \times 1$ convolution with a stride of $2$ to reduce the spatial dimensions and adjust the number of channels, and normalizes it in batches. The main chain processes the input through two convolutional layers. The first layer applies a convolution with a kernel size of $3 \times 3$, stride of $2$, and padding of $1$, which simultaneously reduces the spatial dimensions by half and increases the number of channels. Batch normalization layer and ReLU activation layer follow this layer to stabilize training and introduce non-linearity. The second convolutional layer maintains the same number of channels and spatial dimensions, further refining the learned features before applying another batch normalization. The result from the side chain and the transformed input are summed before the last ReLU activation layer.

We have further differentiated the models based on the inclusion of PI, resulting in four variants: VggOnion, VggOnion-PI, ResOnion, and ResOnion-PI. Models equipped with PI will introduce the extracted features of PI before the output representation. Additionally, we tested the required execution time (ET) for a single sample, obtained by averaging the time taken from 1000 individual samples, on a desktop computer equipped with a 12th Gen Intel(R) Core(TM) i7-12700 CPU @ 2.10 GHz and accelerated by an NVIDIA GPU, specifically the GeForce RTX 3050 OEM version, and compared it with F-B code and NSGPT code, as shown in **Table 2**. For reconstruction on the CPU, these surrogate models are at least 100 times faster than the traditional methods, and for reconstruction on the GPU, they only take a few milliseconds, which is more than 700 times faster than traditional methods. This demonstrates the substantial time efficiency advantage of the surrogate models over F-B code and NSGPT code, along with their potential for application in real-time control and decision-making in tokamak reactors.

**Table 2** Execution time by different methods on both CPU and GPU.

| Method | VggOnion | VggOnion-PI | ResOnion | ResOnion-PI | F-B code | NSGPT code |
|---|---|---|---|---|---|---|
| HL-2A-ET@GPU (ms) | 2.7 | 1.3 | 2.2 | 5.2 | n/a | n/a |
| HL-2A-ET@CPU (ms) | 5.5 | 8.7 | 7.4 | 10.3 | n/a | ≥3700** |
| EAST-ET@GPU (ms) | 6.3 | 7.7 | 9.3 | 8.9 | n/a | n/a |
| EAST-ET@CPU (ms) | 35.7 | 55.7 | 67.3 | 58 | ≥7000* | n/a |

* This value is cited from the reference[26], calculated on a laptop with an Intel(R) Core(TM) i7-8750H CPU @ 2.20 GHz and accelerated by an NVIDIA GPU GTX 1060; ** This value is cited from the reference[25], calculated on A100 GPU with an Intel Xeon Platinum 8352V CPU @ 2.10 GHz.

# 4 Results

This section focuses on the rationale behind the adoption of the incorporation of PI, the Softplus activation function, and the introduction of additional loss terms. We employ two metrics to evaluate the performance of our models. The first metric is the average relative error between the reconstruction profiles and the target profiles, denoted as $E_1$.

$$E_1 = \frac{1}{m}\sum_{j=1}^{m}\left(\frac{1}{r*z}\sum\left|\frac{Net(x)^j - y^j}{y^j_{max}}\right|\right) \quad 4\text{-}1$$

$m$ is the number of test set. $Net(x)^j$ is the reconstruction profile of j-th input $x$. $y^j$ is the target profile of j-th sample and $y^j_{max}$ is the maximum value of $y^j$. The second metric is the average relative error between the BPs and the input data, denoted as $E_2$.

$$E_2 = \frac{1}{m}\sum_{j=1}^{m}\left(\frac{1}{n}\sum_{i=1}^{n}\left|\frac{x_i^j - R_i \cdot Net(x)^j}{x_{max}^j}\right|\right) \quad 4\text{-}2$$

$n$ is the number of LOSs of the j-th input $x$. $x_i^j$ is the i-th LOS of $x^j$. $R_i \cdot Net(x)^j$ is the BP of i-th LOS with $Net(x)^j$, and $x_{max}^j$ is the maximum value of $x^j$. $E_1$ characterizes the model's fitting capability to the target profiles; a lower $E_1$ signifies that the model is more aligned with the inversion algorithms used to generate the target profiles. $E_2$ represents the degree to which the model's results conform to physical principles; a lower $E_2$ indicates that the model's predictions are more consistent with the underlying laws of physics. Correspondingly, for the $j$-th sample, we use relative error $\varepsilon_1$ in Equation (4-3) to evaluate the performance of the model on the single test sample. It is noteworthy that the global maximum values $x_{max}^k$, $y_{max}^j$ and $x_{max}^j$ are chosen as the denominators in in Equation (2-2), Equation (4-1) and Equation (4-2) to prevent large relative errors in the calculation when the denominator is small according to the work[42].

$$\varepsilon_1 = \left|\frac{Net(x)^j\text{-}y^j}{y_{max}^j}\right| \quad 4\text{-}3$$

## 4.1 Role of physical information

To examine the influence of integrating PI on model performance, this section focuses on comparing two sets of models: VggOnion with VggOnion_PI, and ResOnion with ResOnion_PI. The comparisons are designed to elucidate how PI contributes to improving predictive accuracy. For this study, we exclusively use ReLU activation functions in the last two FC layers of each model and the models' loss function includes only one term, $loss_1$, so that evaluates the contributions of PI from side chains rather than PILF. The performance of four models across four datasets is listed in the following **Table 3**.

**Table 3** The impact of integrating PI on model performance. VggOnion, VggOnion_PI, ResOnion and ResOnion_PI are applied on both synthetic datasets and experimental datasets. $E_1$ and $E_2$ are employed to assess models.

| Dataset | Synthetic_EAST ($\bar{\varepsilon} = 3.70 \times 10^{-8}$) | | | |
|---|---|---|---|---|
| Model | VggOnion | VggOnion_PI | ResOnion | ResOnion_PI |
| $E_1$ | $1.02 \times 10^{-2}$ | $\mathbf{0.36 \times 10^{-2}}$ | $2.63 \times 10^{-2}$ | $\mathbf{0.52 \times 10^{-2}}$ |
| $E_2$ | $1.81 \times 10^{-2}$ | $\mathbf{0.56 \times 10^{-2}}$ | $4.88 \times 10^{-2}$ | $\mathbf{0.83 \times 10^{-2}}$ |
| Dataset | Synthetic_HL-2A ($\bar{\varepsilon} = 2.54 \times 10^{-8}$) | | | |
| Model | VggOnion | VggOnion_PI | ResOnion | ResOnion_PI |
| $E_1$ | $0.91 \times 10^{-2}$ | $\mathbf{0.53 \times 10^{-2}}$ | $0.95 \times 10^{-2}$ | $\mathbf{0.75 \times 10^{-2}}$ |
| $E_2$ | $1.36 \times 10^{-2}$ | $\mathbf{0.70 \times 10^{-2}}$ | $1.35 \times 10^{-2}$ | $\mathbf{1.03 \times 10^{-2}}$ |
| Dataset | Exp_EAST ($\bar{\varepsilon} = 4.96 \times 10^{-2}$) | | | |
| Model | VggOnion | VggOnion_PI | ResOnion | ResOnion_PI |
| $E_1$ | $0.60 \times 10^{-2}$ | $\mathbf{0.57 \times 10^{-2}}$ | $0.65 \times 10^{-2}$ | $\mathbf{0.57 \times 10^{-2}}$ |
| $E_2$ | $4.95 \times 10^{-2}$ | $\mathbf{4.76 \times 10^{-2}}$ | $4.90 \times 10^{-2}$ | $\mathbf{4.67 \times 10^{-2}}$ |
| Dataset | Exp_HL-2A ($\bar{\varepsilon} = 5.42 \times 10^{-2}$) | | | |
| Model | VggOnion | VggOnion_PI | ResOnion | ResOnion_PI |
| $E_1$ | $0.37 \times 10^{-2}$ | $\mathbf{0.26 \times 10^{-2}}$ | $0.32 \times 10^{-2}$ | $\mathbf{0.28 \times 10^{-2}}$ |
| $E_2$ | $5.43 \times 10^{-2}$ | $\mathbf{5.42 \times 10^{-2}}$ | $5.46 \times 10^{-2}$ | $\mathbf{5.42 \times 10^{-2}}$ |

For datasets with almost zero error $\bar{\varepsilon}$ (Synthetic_EAST and Synthetic_HL-2A), the introduction of PI has an improvement on model performance. Despite the overall metrics being relatively small due to normalization, the inclusion of PI offers a relative enhancement compared to models that do not incorporate PI. Both $E_1$ and $E_2$ drop, with $E_1$ dropping by $0.84 \times 10^{-2}$ about 52% on average and $E_2$ dropping by $1.57 \times 10^{-2}$ about 56% on average. **Figure 11** shows the target profile of the $j$-th sample in the Synthetic_HL-2A test set and **Figure 12** shows the test results of different models on the $j$-th sample in the Synthetic_HL-2A test set. With the introduction of PI, the $\varepsilon_1$ is reduced, as can be seen by comparing **Figure 12**(b) with **Figure 12**(e), and **Figure 12**(h) with **Figure 12**(k). The BPs of reconstruction profile (green dots) is closer to the inputs (blue triangle) when comparing the magnified views of **Figure 12**(c) and **Figure 12**(f), as well as **Figure 12**(i) and **Figure 12**(l). The reduction in the discrepancy between the BPs of the reconstruction profile (green dots) and inputs (blue triangles) is understandable. This indicates that as the reconstruction profile more closely resembles the target profile, the BPs of the reconstruction profile will progressively approach the BPs of the target profile (red dots), which are themselves very close to the inputs, given that the $\bar{\varepsilon}$ of the synthetic dataset is virtually zero.

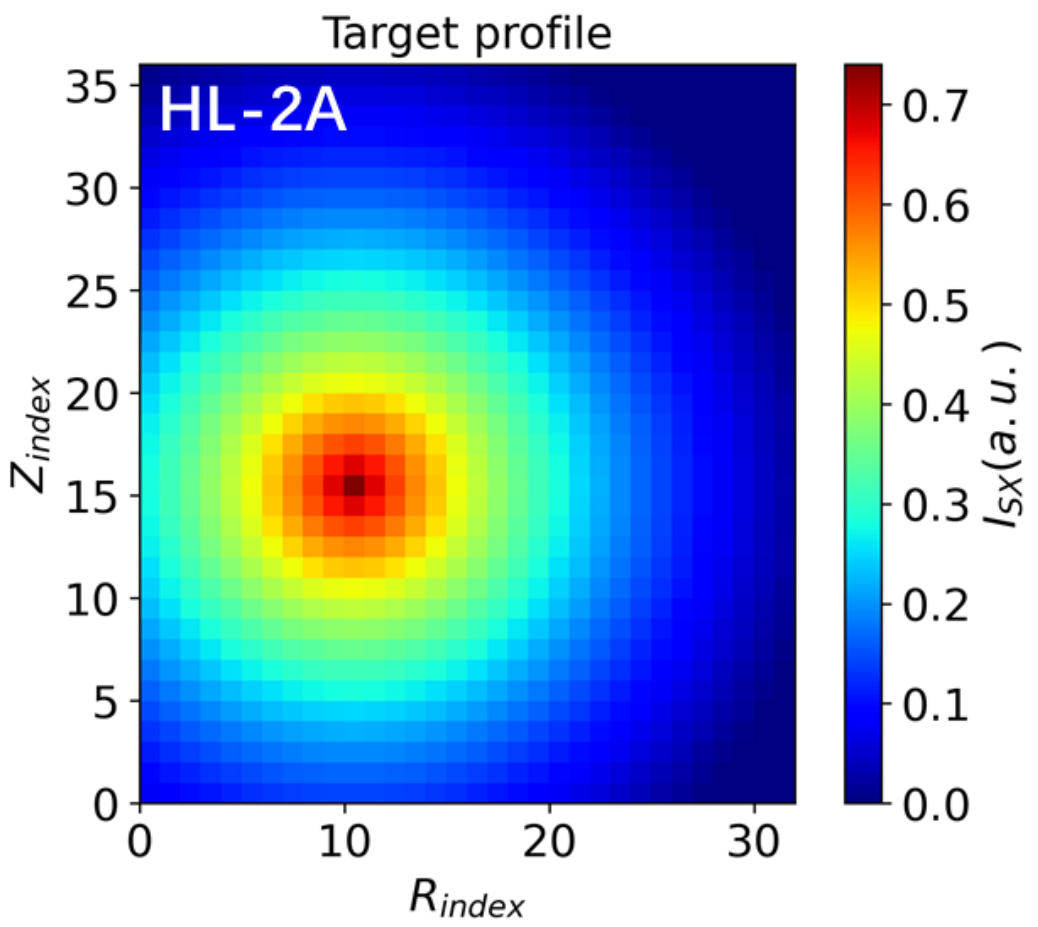

**Figure 11** Target profile of the j-th sample in the Synthetic_HL-2A test set.

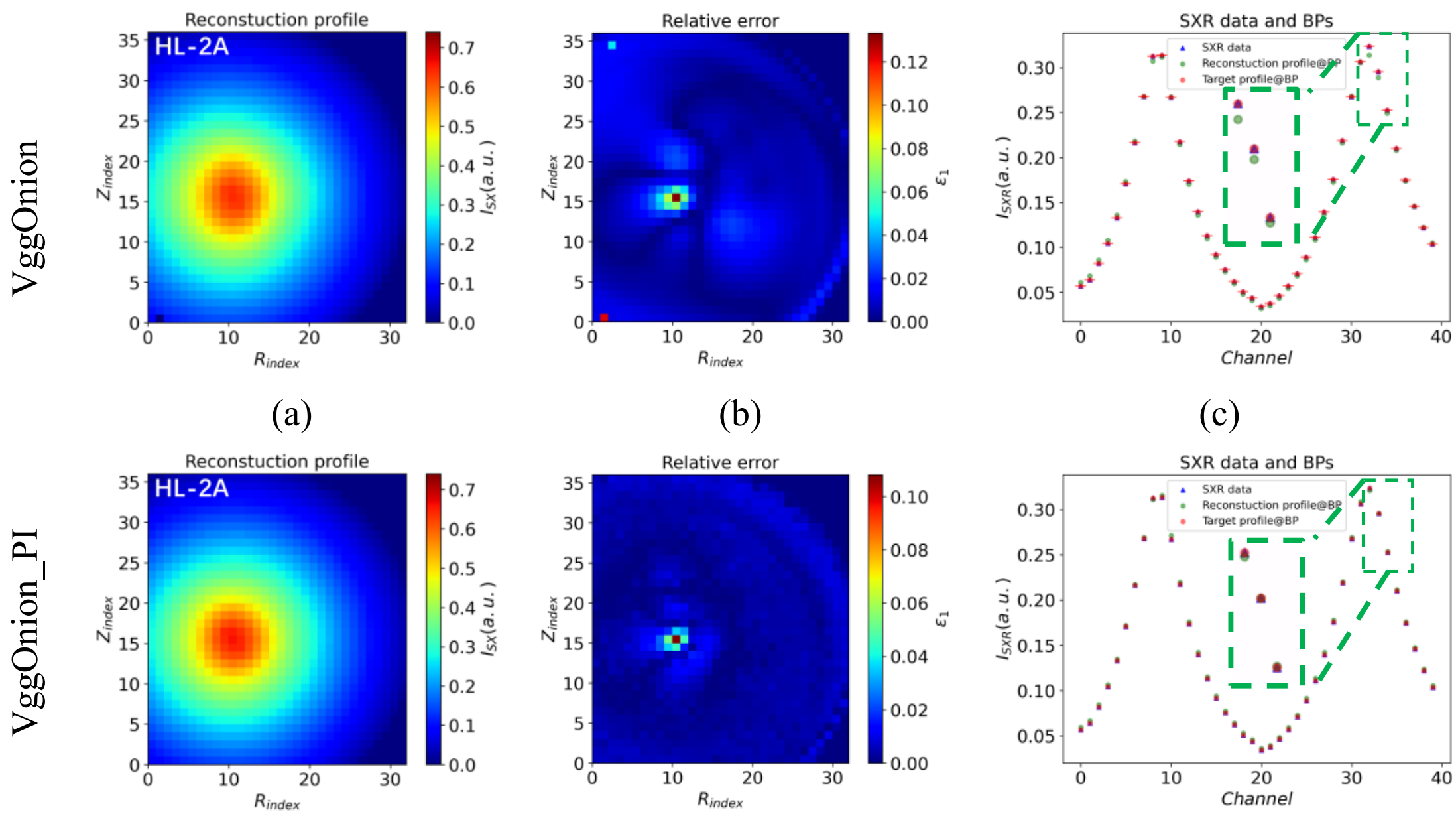

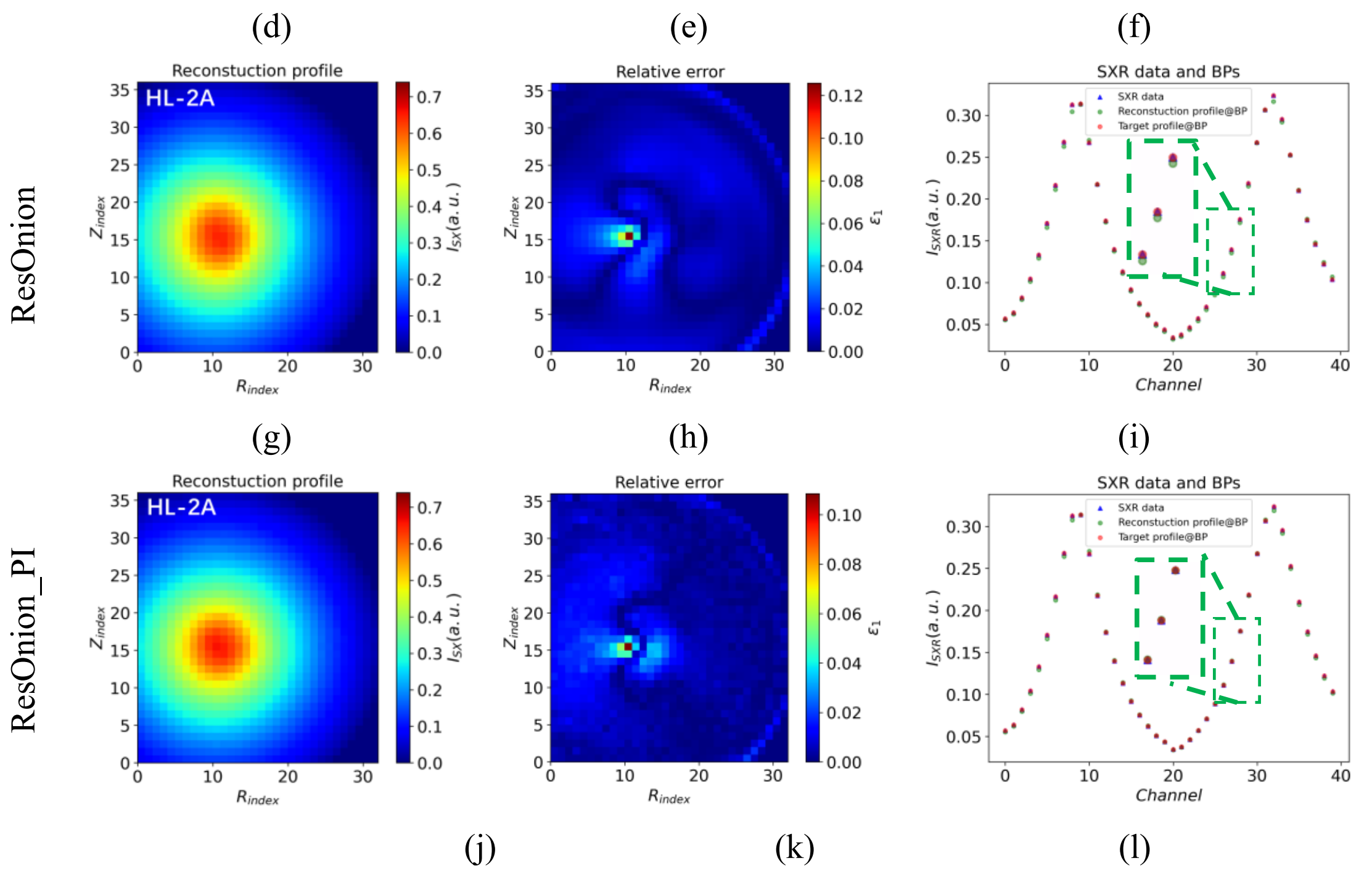

**Figure 12** Test results of different models on the j-th sample in the Synthetic_HL-2A test set. First column: reconstruction profiles for each model. Second column: distributions of $\varepsilon_1$. Third column: comparisons between inputs and BPs for various channels.

For datasets from experiments with error $\bar{\varepsilon}$ around $5 \times 10^{-2}$ (Exp_EAST and Exp_HL-2A), VggOnion_PI and ResOnion_PI do not have demonstrated commendable performance. $E_1$ decreased slightly, with an average reduction of $0.06 \times 10^{-2}$ approximately 15%, whereas $E_2$ remained almost unchanged. Despite the BPs of reconstruction profiles being close to those of target profiles indicating that the performance of the surrogate models is comparable to that of the inversion algorithms, due to the presence of $\bar{\varepsilon}$, the discrepancy between the BPs of reconstruction profiles and inputs has not been reduced. **Figure 13** illustrates the target profile of the j-th sample in the Exp_HL-2A test set, while **Figure 14** presents the test results of different models on the j-th sample in the Exp_HL-2A test set. The performance of the four models is comparable. By comparing the results from the synthetic datasets with those from the experimental datasets, it can be inferred that the noise introduced by inherent error $\bar{\varepsilon}$ of datasets may limit the extent of performance improvements achieved through the incorporation of PI. This may be due to the presence of noise, which makes it difficult for the model to distinguish between useful information and irrelevant disturbances. Furthermore, this issue can render the PI insufficient for guiding the model to correctly focus on the key features within the data.

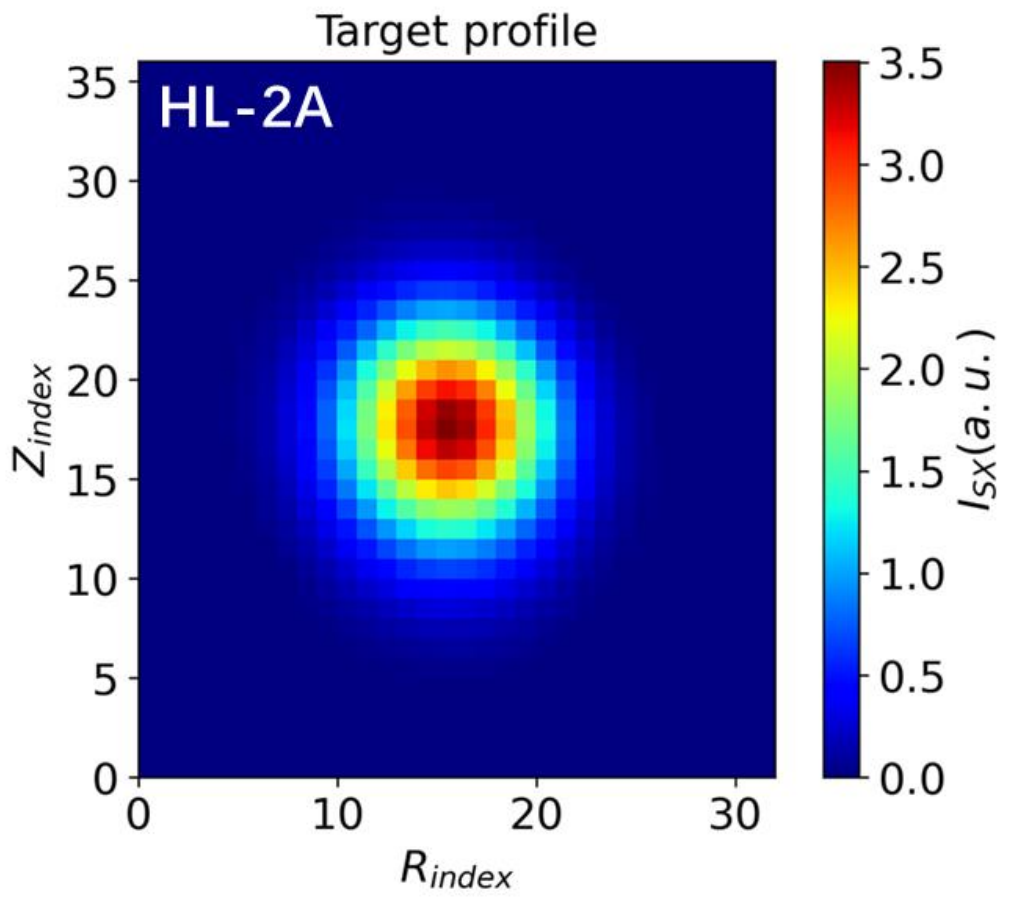

**Figure 13** Target profile of the j-th sample in the Exp_HL-2A test set.

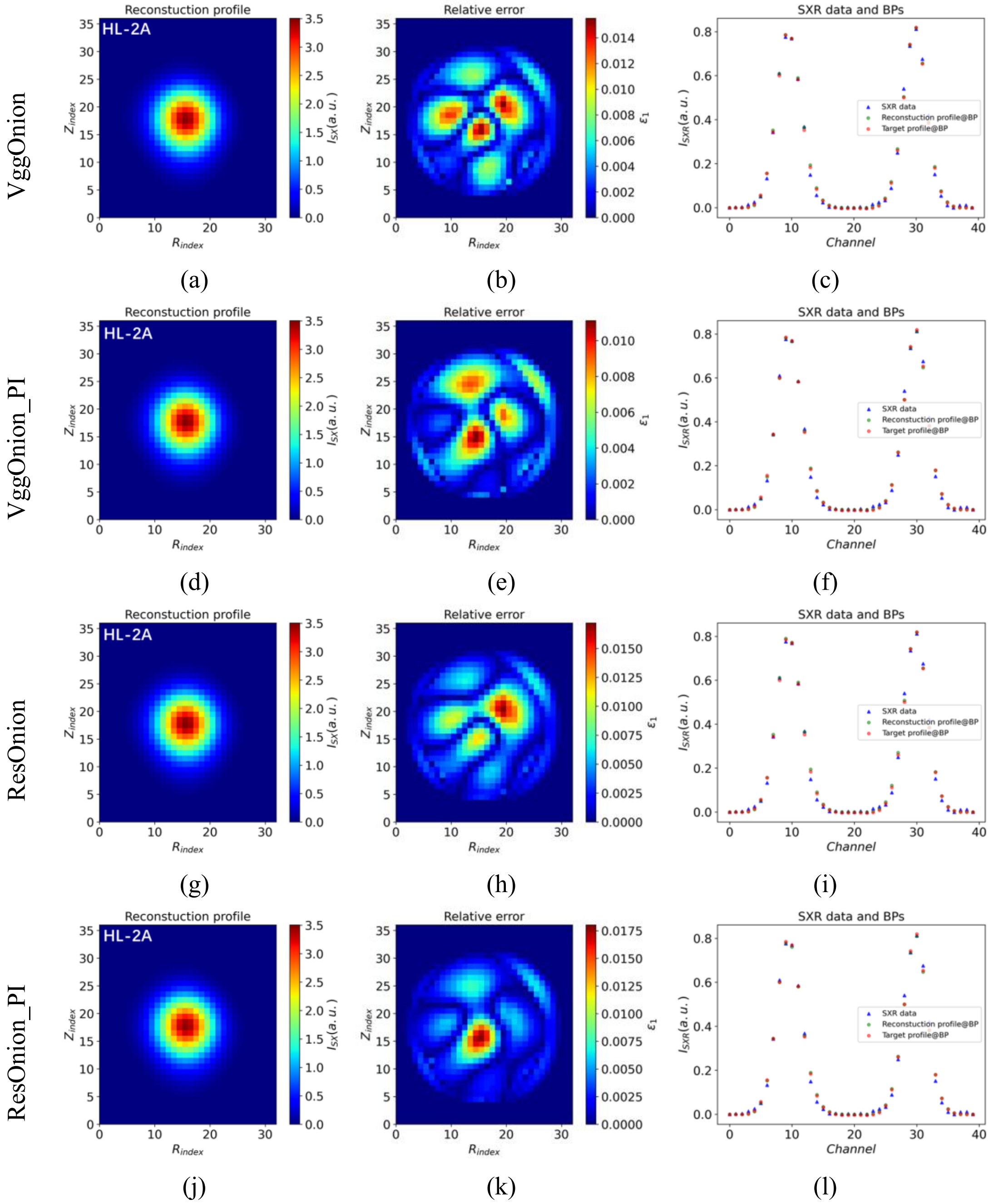

**Figure 14** Test results of different models on the j-th sample in the Exp_HL-2A test set. First column: reconstruction profiles for each model. Second column: distributions of $\varepsilon_1$. Third column: comparisons between inputs and BPs for various channels.

## 4.2 Role of the Softplus activation function

In the results presented in the previous section, we observed implausible values at the edges of the predicted result contour plots, i.e., the edges are not smooth, shown in the localized magnified grayscale images in **Figure 16**(a) and **Figure 16**(c). Smooth edges reflect the continuity of the reconstructed profile, whereas jagged edges suggest the presence of steps in the reconstructed profile, which is inconsistent with physical reality. This section primarily focuses on addressing this issue. We think it comes from the characteristics of the ReLU activation function as mentioned in **Section 4.1**. After comparing several activation functions, we find the Softplus activation function which can be considered a smoothed version of the ReLU function. This characteristic helps to mitigate the issue of vanishing gradients that can occur with ReLU when $x < 0$, as it prevents the output from becoming zero and thus maintains a more stable gradient flow during training.

We conduct a comparative analysis between the VggOnion and ResOnion models to evaluate the impact of activation functions used in the final two FC layers on model performance. The models' loss function is solely composed of the term $loss_1$. The performance of models with ReLU or Softplus function is shown in the following **Table 4**.

**Table 4** reveals that the adoption of the Softplus function also leads to an enhancement in model performance. For Synthetic_EAST and Synthetic_HL-2A datasets, the introduction of Softplus function has an improvement on two models. Both $E_1$ and $E_2$ drop, with $E_1$ dropping by $1.06 \times 10^{-2}$ about 71% on average and $E_2$ dropping by $1.87 \times 10^{-2}$ about 73% on average. For Exp_EAST and Exp_HL-2A datasets, both $E_1$ and $E_2$ have decreased slightly, with $E_1$ decreasing by $0.11 \times 10^{-2}$ about 27% on average and $E_2$ decreasing by $0.12 \times 10^{-2}$ about 2.3% on average. Once again, the comparison between the synthetic and experimental datasets suggests that the noise resulting from inherent error $\bar{\varepsilon}$ of datasets may also impede the performance enhancements that could be realized by adopting the Softplus activation function.

**Table 4** The impact of the Softplus activation function on model performance. VggOnion and ResOnion are applied on both synthetic datasets and experimental datasets. $E_1$ and $E_2$ are employed to assess models.

| Dataset | Synthetic_EAST ($\bar{\varepsilon} = 3.70 \times 10^{-8}$) | | | |
|---|---|---|---|---|
| Model | VggOnion | | ResOnion | |
| Activation Function | ReLU | Softplus | ReLU | Softplus |
| $E_1$ | $1.02 \times 10^{-2}$ | $\mathbf{0.27 \times 10^{-2}}$ | $2.63 \times 10^{-2}$ | $\mathbf{0.25 \times 10^{-2}}$ |
| $E_2$ | $1.81 \times 10^{-2}$ | $\mathbf{0.41 \times 10^{-2}}$ | $4.88 \times 10^{-2}$ | $\mathbf{0.44 \times 10^{-2}}$ |
| Dataset | Synthetic_HL-2A ($\bar{\varepsilon} = 2.54 \times 10^{-8}$) | | | |
| Model | VggOnion | | ResOnion | |
| Activation Function | ReLU | Softplus | ReLU | Softplus |
| $E_1$ | $0.91 \times 10^{-2}$ | $\mathbf{0.39 \times 10^{-2}}$ | $0.95 \times 10^{-2}$ | $\mathbf{0.36 \times 10^{-2}}$ |
| $E_2$ | $1.36 \times 10^{-2}$ | $\mathbf{0.55 \times 10^{-2}}$ | $1.35 \times 10^{-2}$ | $\mathbf{0.50 \times 10^{-2}}$ |
| Dataset | Exp_EAST ($\bar{\varepsilon} = 4.96 \times 10^{-2}$) | | | |
| Model | VggOnion | | ResOnion | |

| Activation Function | ReLU | Softplus | ReLU | Softplus |
|---|---|---|---|---|
| $E_1$ | $0.60 \times 10^{-2}$ | $\mathbf{0.51 \times 10^{-2}}$ | $0.65 \times 10^{-2}$ | $\mathbf{0.59 \times 10^{-2}}$ |
| $E_2$ | $4.95 \times 10^{-2}$ | $\mathbf{4.82 \times 10^{-2}}$ | $4.90 \times 10^{-2}$ | $\mathbf{4.69 \times 10^{-2}}$ |
| Dataset | Exp_HL-2A ($\bar{\varepsilon} = 5.42 \times 10^{-2}$) | | | |
| Model | VggOnion | | ResOnion | |
| Activation Function | ReLU | Softplus | ReLU | Softplus |
| $E_1$ | $0.37 \times 10^{-2}$ | $\mathbf{0.20 \times 10^{-2}}$ | $0.32 \times 10^{-2}$ | $\mathbf{0.20 \times 10^{-2}}$ |
| $E_2$ | $5.43 \times 10^{-2}$ | $\mathbf{5.37 \times 10^{-2}}$ | $5.46 \times 10^{-2}$ | $\mathbf{5.39 \times 10^{-2}}$ |

**Figure 15** shows the target profile of the sample in the Exp_EAST test set and Figure 16 compares the performance of the VggOnion and ResOnion models when the activation functions of the final two FC layers are set to ReLU and Softplus, respectively. Upon examining the magnified portions of **Figure 16**(b) and **Figure 16**(d), it is evident that the results obtained with the Softplus activation function exhibit smoother edges and are more closely aligned with the target profiles. Besides, we can see the $\varepsilon_1$ at the edge of reconstruction profiles has decreased from about 0.03 to about 0.005 by comparing **Figure 17**(a) with **Figure 17**(b), and **Figure 17**(c) with **Figure 17**(d). This is attributed to the continuous nature of the Softplus function for $x < 0$ and its property of yielding positive output values, which aligns more closely with the characteristics of the underlying physical quantities.

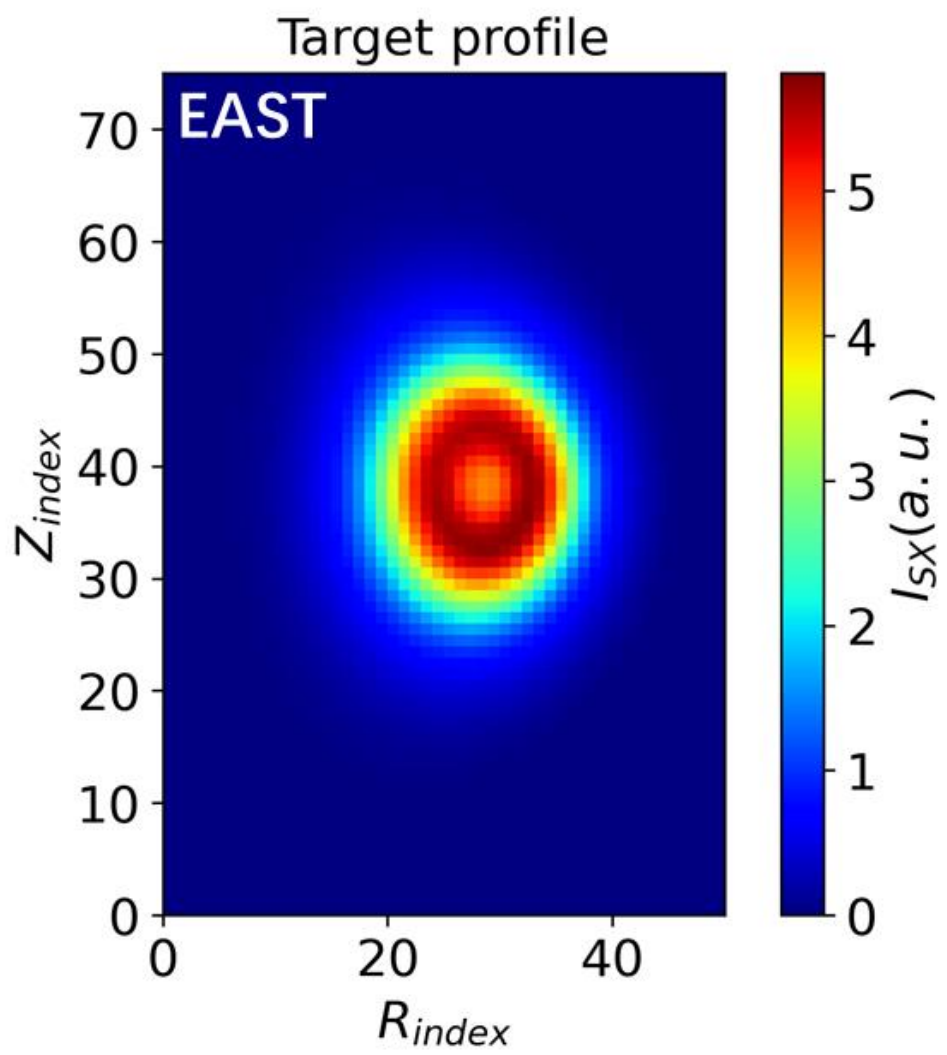

**Figure 15** The target profile of the j-th sample in the Exp_EAST test set.

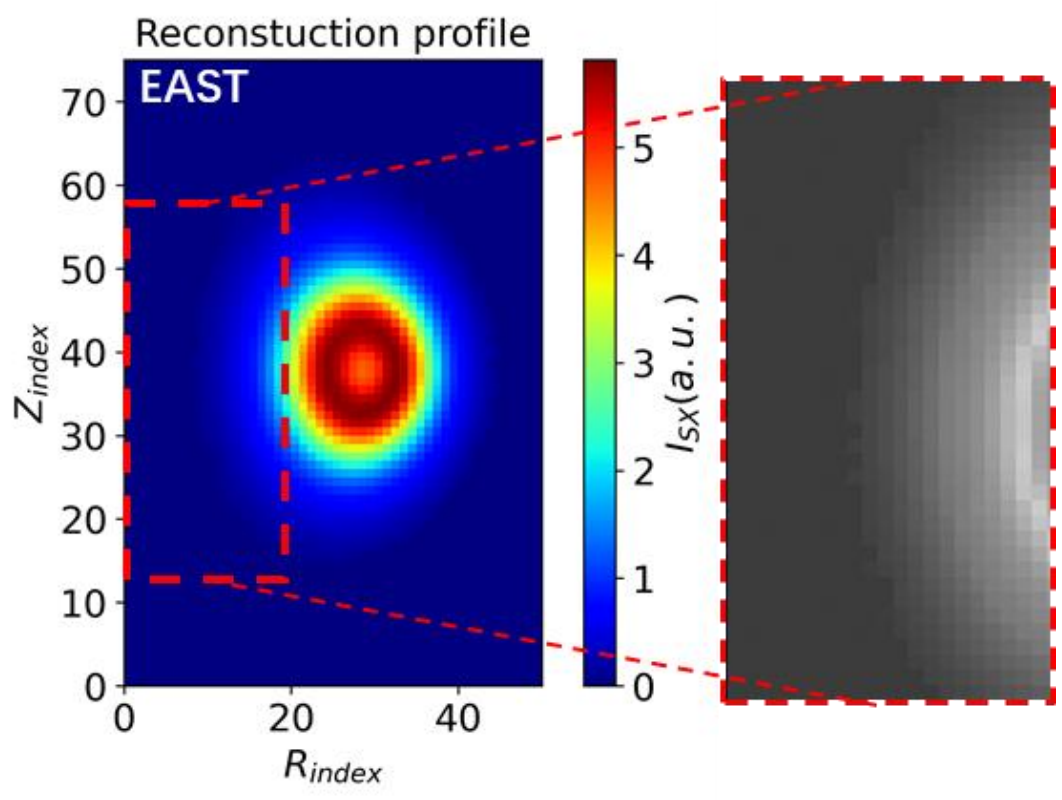

(a) VggOnion with ReLU

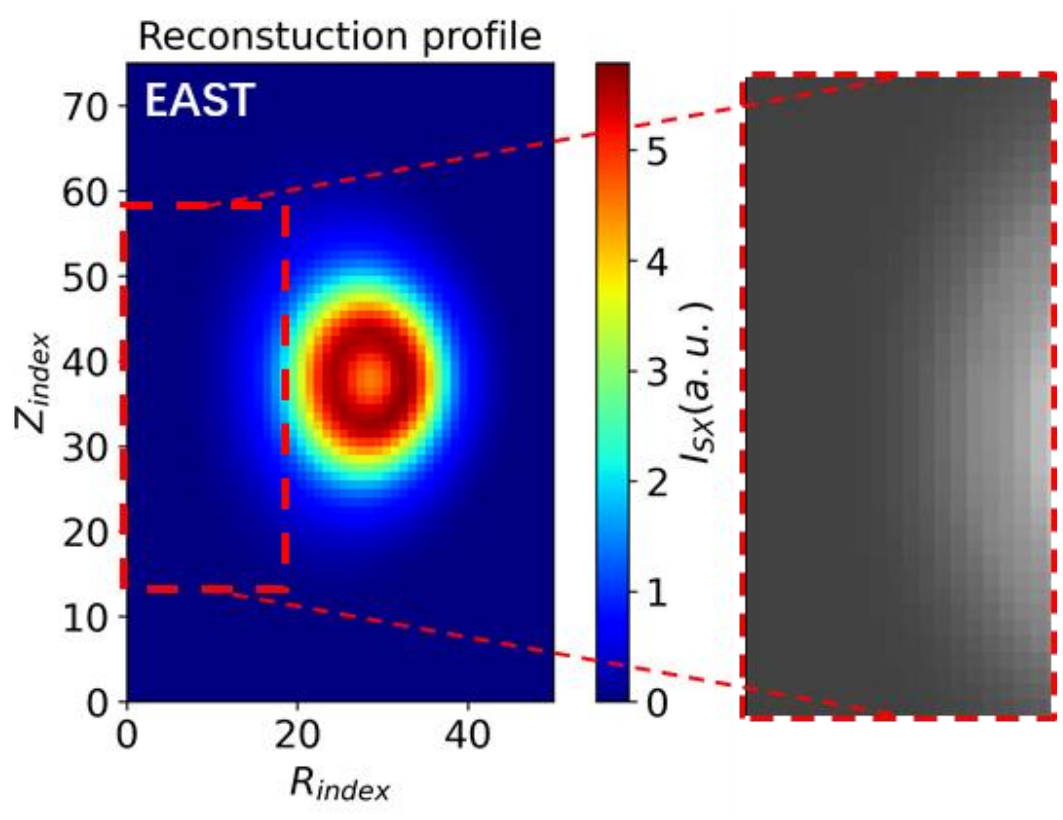

(b) VggOnion with Softplus

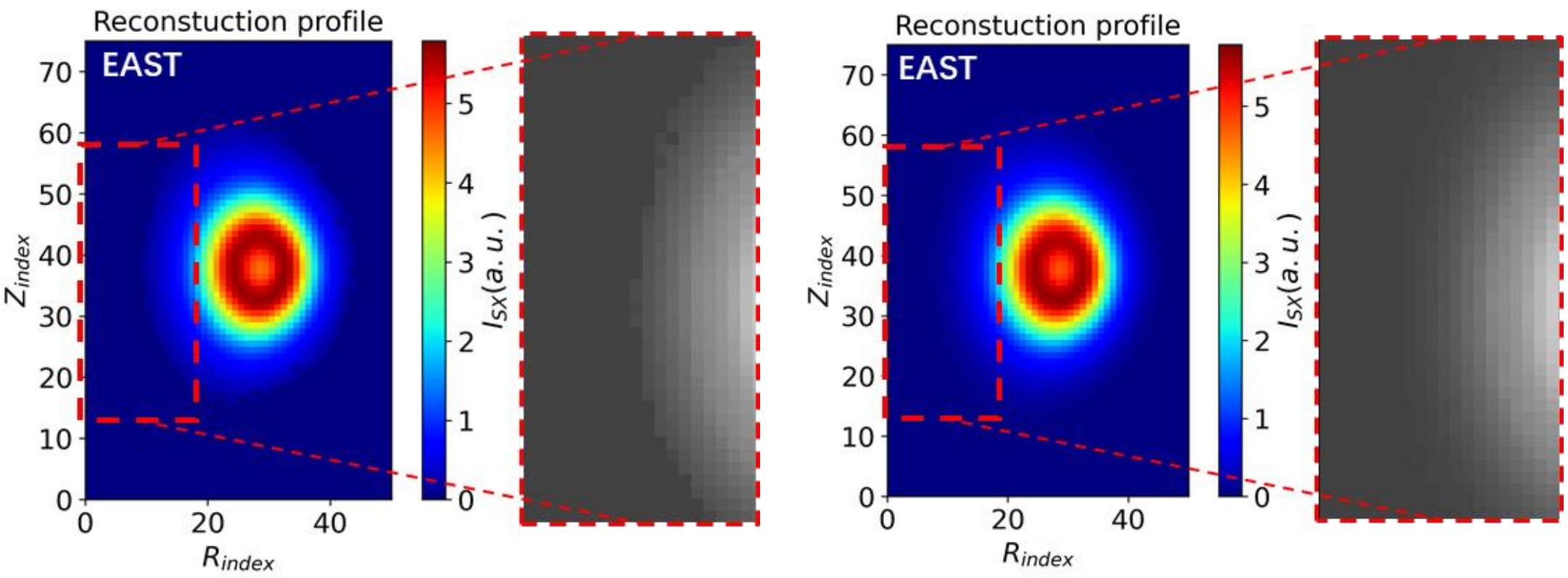

(c) ResOnion with ReLU (d) ResOnion with Softplus

**Figure 16** Test results of different models on the j-th sample in the Exp_EAST test set: (a) Reconstruction profile from VggOnion with ReLU activation function; (b) Reconstruction profile from VggOnion with Softplus activation function; (c) Reconstruction profile from ResOnion with ReLU activation function; (d) Reconstruction profile from ResOnion with Softplus activation function.

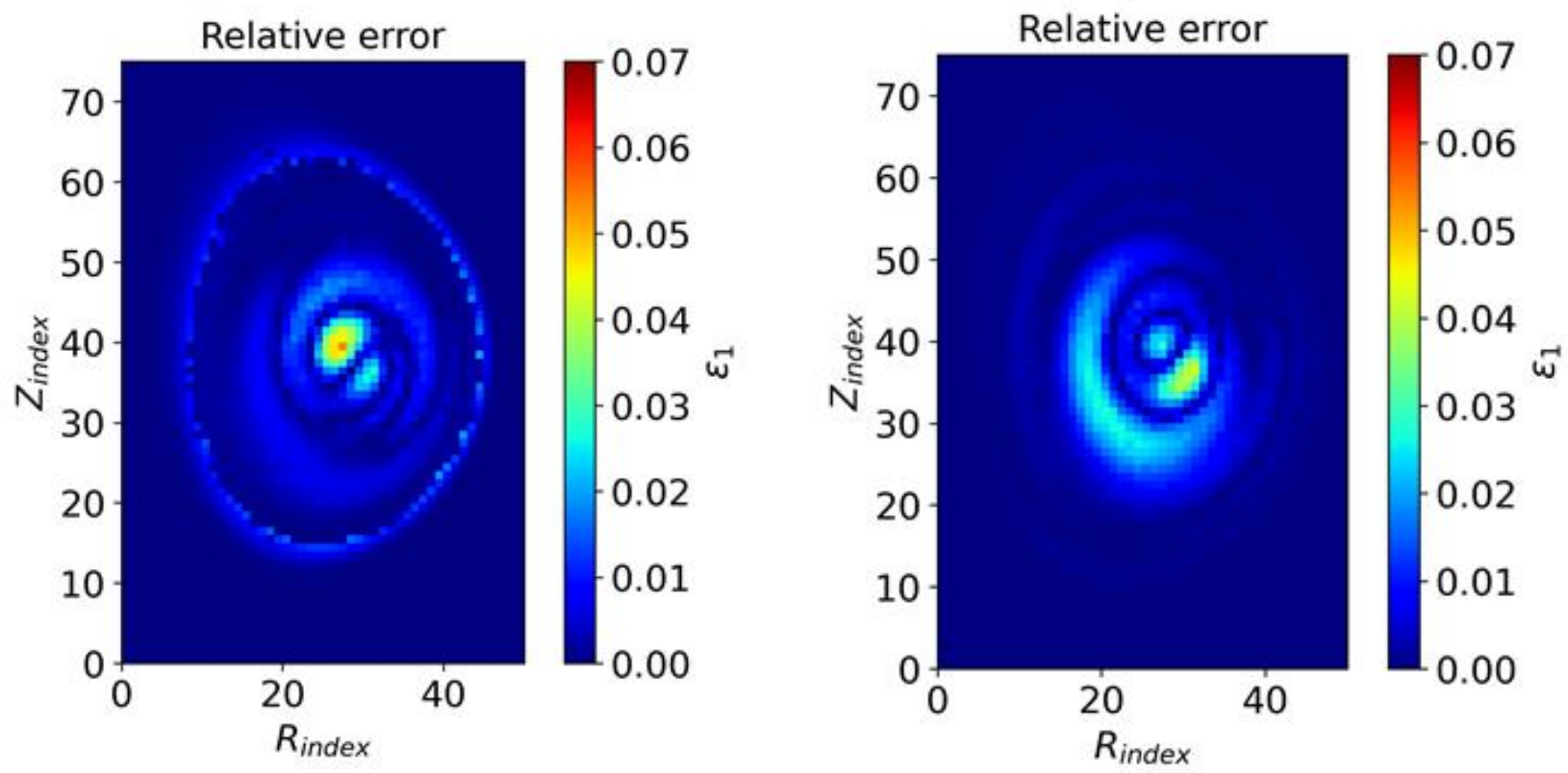

(a) VggOnion with ReLU (b) VggOnion with Softplus

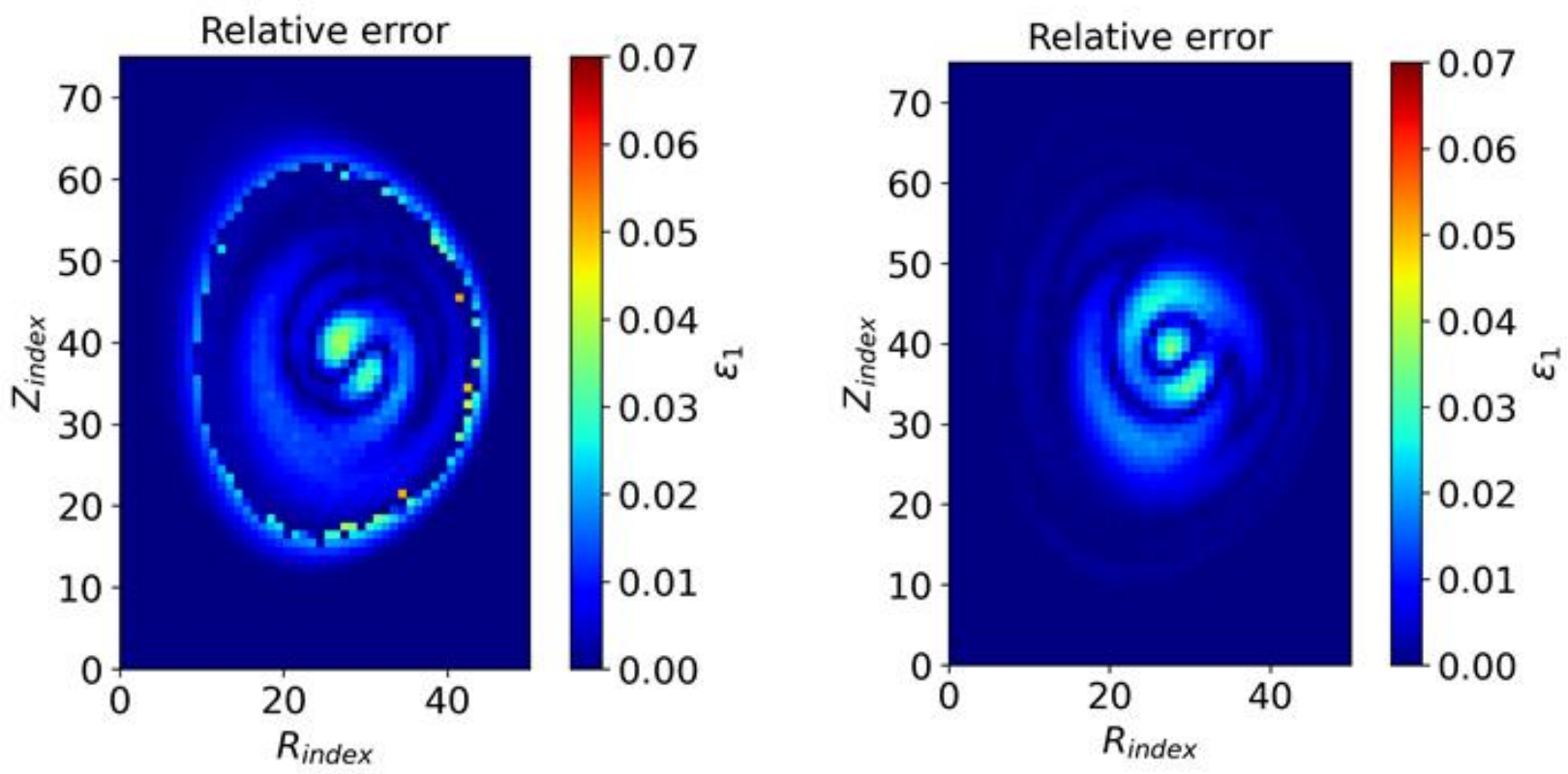

(c) ResOnion with ReLU (d) ResOnion with Softplus

**Figure 17** Distributions of $\varepsilon_1$ of different models on the j-th sample in the Exp_EAST test set: (a) Distribution of $\varepsilon_1$ from VggOnion with ReLU activation function; (b) Distribution of $\varepsilon_1$ from VggOnion with Softplus activation function; (c) Distribution of $\varepsilon_1$ from ResOnion with ReLU activation function; (d) Distribution of $\varepsilon_1$ from ResOnion with Softplus activation function.

## 4.3 Role of the physics-informed loss function

Building upon the findings from the previous section, this section is dedicated to assessing the impact of the PILF on model performance. The models utilized in this section are the VggOnion_PI and ResOnion_PI. Softplus function is used in the final two FC layers. The additional loss terms in the PILF are designed to further refine the model's predictions, ensuring that not only reconstruction profiles fit the target profiles but also BPs of reconstruction profiles close to the diagnostic data.

For the error-free synthetic dataset, closer model predictions to the target profiles imply that the BPs will also be closer to the input data. In such cases, utilizing the term $loss_1$ is enough. However, for experimental datasets with inherent errors, a close match between model predictions and target profiles does not guarantee the closeness of the BPs to the input. It is understood that when BP closely aligns with the input, it indicates that the predictions better satisfy the inherent experimental constraints. This alignment, in turn, may lead to a deviation between model predictions and target profiles for experimental datasets with inherent errors. Therefore, a hyperparameter in PILF is employed to achieve a balanced optimization. This section adopts experimental datasets for model training to account for these complexities and only $E_2$ is considered to assess the model. The hyperparameter $\mathrm{c}_1$ of the $loss_2$ must be adjusted according to the model and the dataset to reflect the desired emphasis on inherent physical constraints during model training. In this section, we illustrate the effects brought about by the hyperparameter $\mathrm{c}_1$ using examples from two different datasets: Exp_EAST and Exp_HL-2A. For the Exp_EAST dataset, the hyperparameter $\mathrm{c}_1$ of the loss function $loss_2$ is set to 0.618, whereas for the Exp_HL-2A dataset, it is set to 1.0.

**Table 5** presents the performance of the VggOnion_PI and ResOnion_PI models on the Exp_EAST and Exp_HL-2A datasets. When compared to the models trained with only $loss_1$, the introduction of the PILF resulted in a decrease in $E_2$ for different models, with the extent of this reduction being related to the hyperparameter $\mathrm{c}_1$. For the Exp_EAST dataset with $\mathrm{c}_1$ set to 0.618, after applying PILF, the average decrease in $E_2$ is $0.95 \times 10^{-2}$ about 19.6%. For the Exp_HL-2A dataset with $\mathrm{c}_1$ set to 1.0, the application of PILF led to a more substantial average reduction in $E_2$, amounting to $4.59 \times 10^{-2}$ about 85.4%. This indicates that PILF can effectively enhance the models' ability to adhere to the inherent experimental constraints in their predictions, and the extent of this enhancement is associated with the hyperparameter $\mathrm{c}_1$.

**Table 5** The impact of the PILF on model performance. VggOnion_PI and ResOnion_PI are applied on experimental datasets, Exp_EAST and Exp_HL-2A. $E_2$ is employed to assess models.

| Dataset | Exp_EAST ($\bar{\varepsilon} = 4.96 \times 10^{-2}$) | | | |
|---|---|---|---|---|
| Model | VggOnion_PI | | ResOnion_PI | |
| Loss Function | $loss_1$ | PILF $\mathrm{c}_1 = 0.618$ | $loss_1$ | PILF $\mathrm{c}_1 = 0.618$ |
| $E_1$ | $0.50 \times 10^{-2}$ | $1.35 \times 10^{-2}$ | $0.45 \times 10^{-2}$ | $1.37 \times 10^{-2}$ |
| $E_2$ | $5.08 \times 10^{-2}$ | $\mathbf{4.42 \times 10^{-2}}$ | $4.69 \times 10^{-2}$ | $\mathbf{3.46 \times 10^{-2}}$ |
| Dataset | Exp_HL-2A ($\bar{\varepsilon} = 5.42 \times 10^{-2}$) | | | |
| Model | VggOnion_PI | | ResOnion_PI | |
| Loss Function | $loss_1$ | PILF $\mathrm{c}_1 = 1.0$ | $loss_1$ | PILF $\mathrm{c}_1 = 1.0$ |
| $E_1$ | $0.21 \times 10^{-2}$ | $1.94 \times 10^{-2}$ | $0.21 \times 10^{-2}$ | $1.93 \times 10^{-2}$ |
| $E_2$ | $5.38 \times 10^{-2}$ | $\mathbf{0.76 \times 10^{-2}}$ | $5.37 \times 10^{-2}$ | $\mathbf{0.82 \times 10^{-2}}$ |

**Figure 18** shows the target profile and **Figure 19** illustrates the performance of the model on the sample with MHD instability in Exp_EAST test dataset. By comparing **Figure 19**(a), **Figure**

**19**(g) and the target profile, we can find that the reconstruction profiles obtained by models trained solely with $loss_1$ cannot well reflect the existence of the magnetic island in the target profile, although the BPs of the reconstruction profiles and the BPs of the target profiles (green and red dots in **Figure 19**(c) and **Figure 19**(i)) are in good agreement. This may be attributed to the fact that the PI introduced by the model only incorporates line integration and lacks MHD-related knowledge. Consequently, the model is unable to capture the characteristics of magnetic islands present in the data, leading to reconstructed profiles that are expected to align more closely with a Gaussian-like distribution.

Models incorporating the PILF show a trend where the BPs of the reconstruction profiles deviate from the BPs of the target profiles and move closer to the inputs in the magnified views of **Figure 19**(f) and **Figure 19**(l). Due to the hyperparameter $c_1$ of the loss function $loss_2$ being set to 0.618, the BPs of the reconstruction profiles do not align perfectly with the input. The **Figure 19**(d) and **Figure 19**(j) reveal that the models with PILF exhibit a noticeable divergence from the target profile (**Figure 18**) in the plasma core region that the in-out asymmetry observed in the reconstruction profiles seems to be in the opposite direction with respect to the target profile. This is because the target profile obtained through inversion algorithm based on experimental inputs indeed contains errors, which is proved by the discrepancies between the BPs of the target profile and the experimental inputs in **Figure 19**(c) and **Figure 19**(i). The target profile fails to capture the shape of the local emissivity in the center, whereas the introduction of PILF can address this limitation. Specifically, it results in the BPs of the reconstructed profile being more aligned with the experimental inputs. Therefore, incorporating PILF can reduce the errors associated with the inversion algorithm, leading to a more accurate reconstruction.

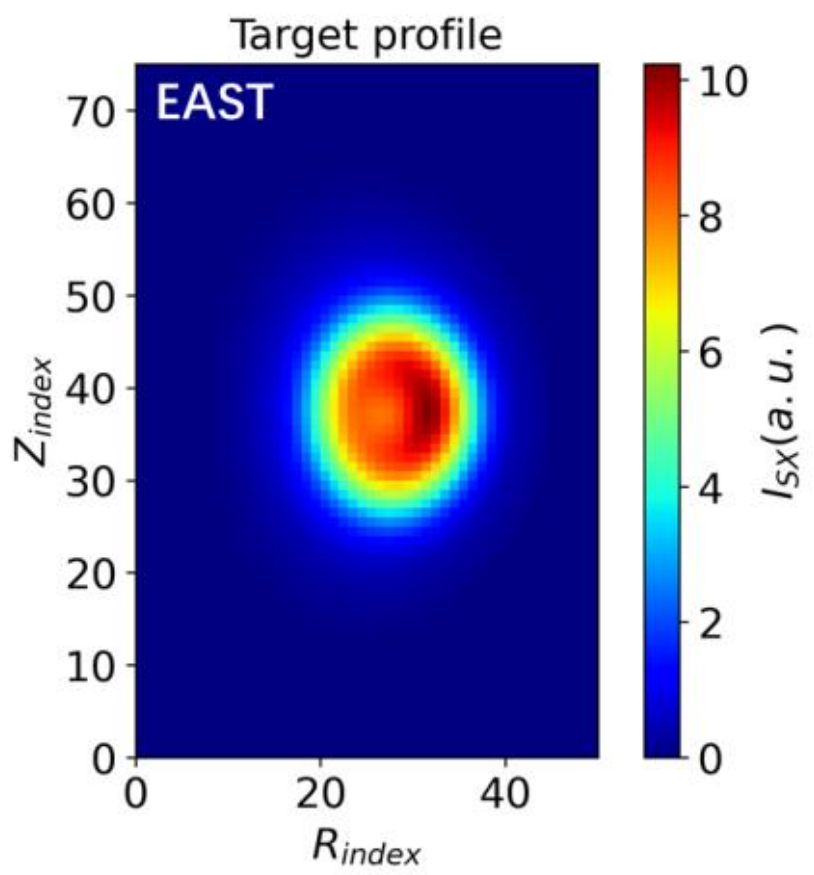

**Figure 18** Target profile of the j-th sample in the Exp_EAST test set.

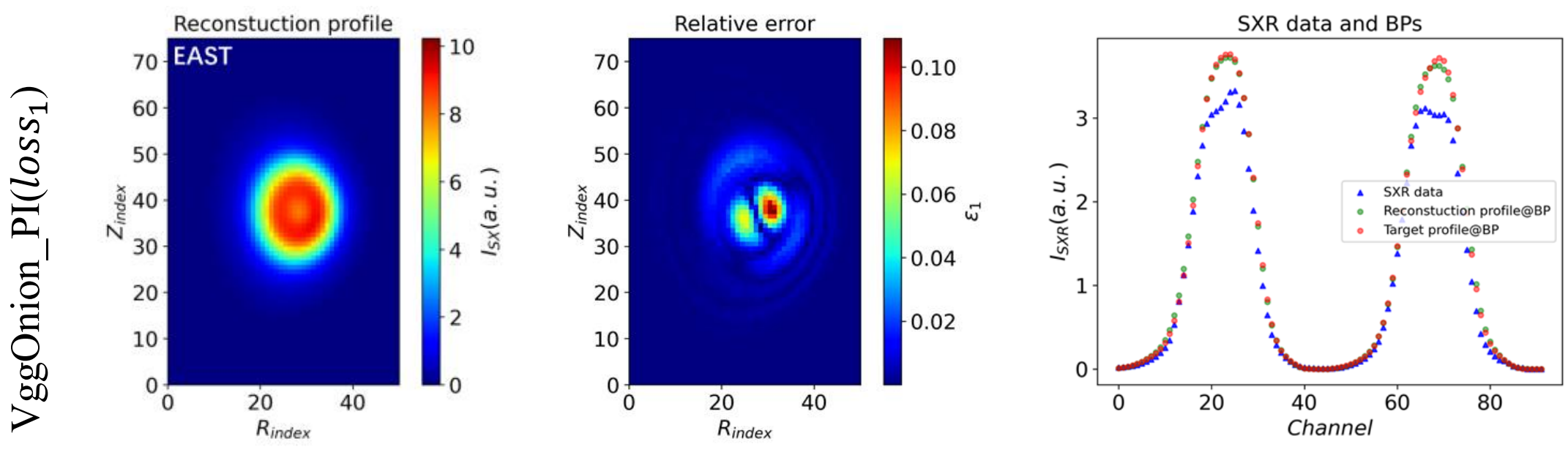

(a) (b) (c)

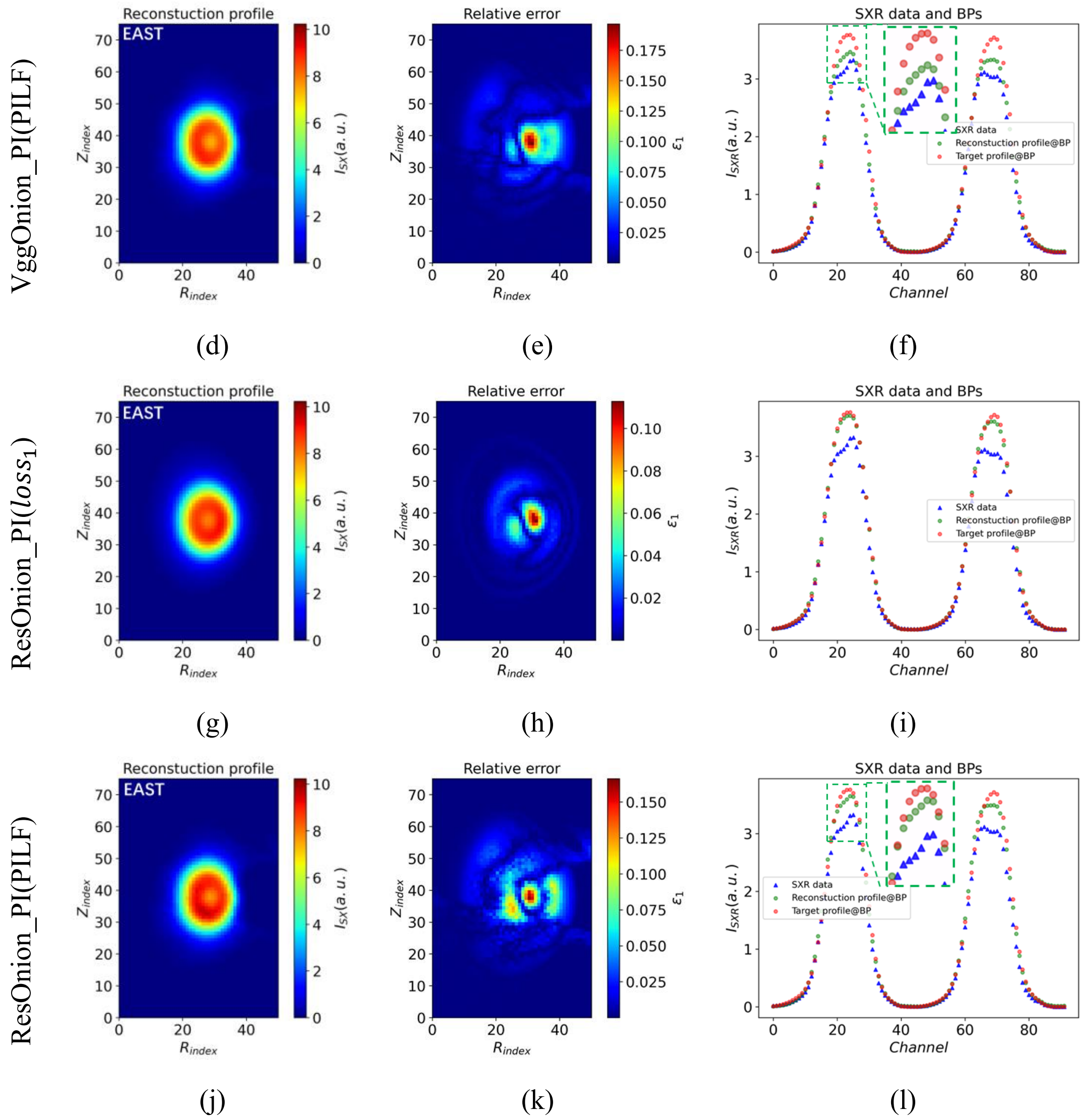

**Figure 19** Test results of different models on the j-th sample in the Exp_EAST test set. First column: reconstruction profiles for each model. Second column: distributions of $\varepsilon_1$. Third column: comparisons between inputs and BPs for various channels.

**Figure 20** shows the target profile and **Figure 21** demonstrates the performance of the model on the Exp_HL-2A test dataset. By comparing the magnified views of **Figure 21**(c) with **Figure 21**(f) and **Figure 21**(i) with **Figure 21**(l), it can be concluded that models incorporating the PILF exhibit BPs of the reconstruction profile (green dots) that are nearly indistinguishable from the input (blue triangles). This near-perfect alignment is attributed to the hyperparameter $c_1$ being set to 1.0, which ensures that the predictions closely match the input. Thus, from the BPs of the reconstruction profile, the accuracy of the surrogate model with PILF is superior to that of the inversion algorithms. However, this approach also introduces some issues; for instance, the reconstruction profiles become less smooth and exhibit localized discontinuities at the edges. To mitigate this phenomenon, additional edge regularization terms could be incorporated into the loss function in future work.

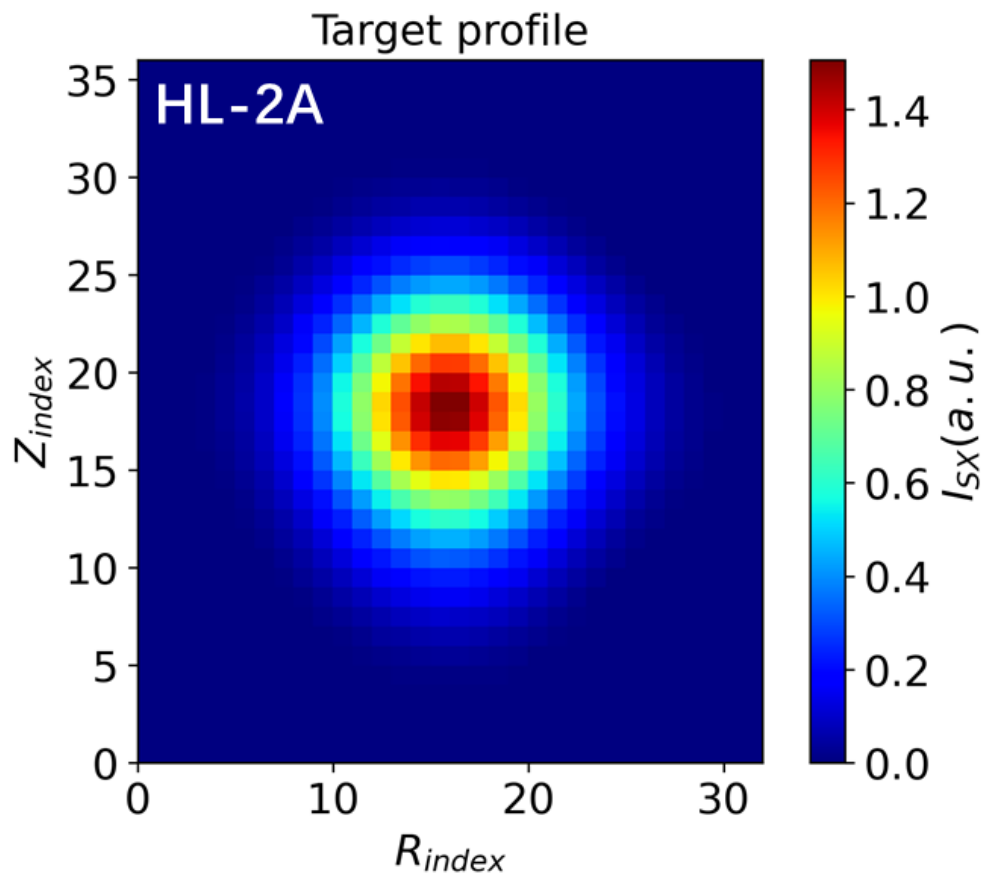

**Figure 20** Target profile of the j-th sample in the Exp_HL-2A test set.

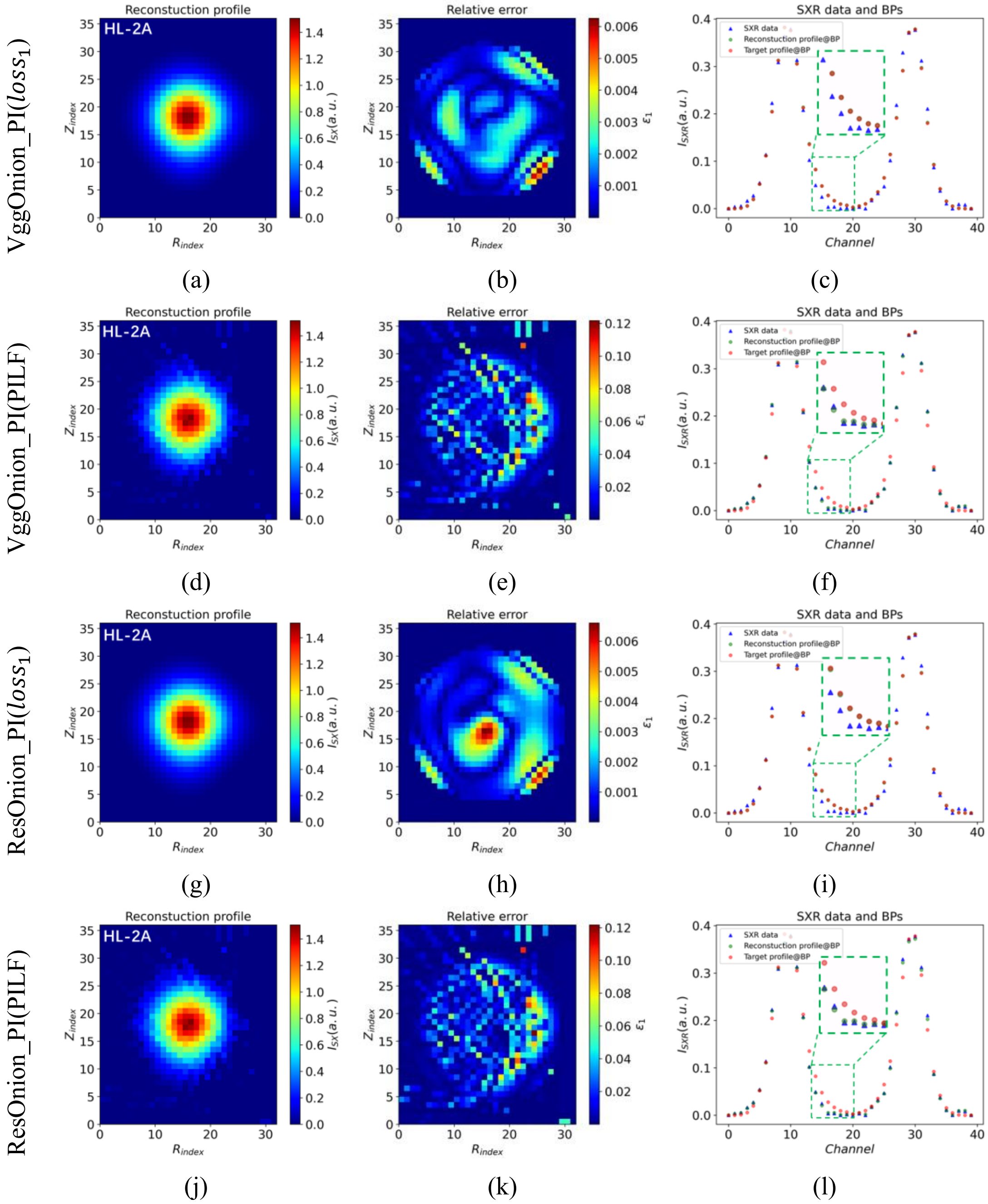

**Figure 21** Test results of different models on the j-th sample in the Exp_HL-2A test set. First column: reconstruction profiles for each model. Second column: distributions of $\varepsilon_1$. Third column: comparisons between inputs and BPs for various channels.

# 5 Conclusion

In this paper, we constructed four models based on the proposed physics-informed deep learning model architecture and conducted experimental analyses using four distinct datasets. All models are examined and show an execution time of several milliseconds on the GPU. This improvement not only accelerates data processing but also supports more efficient experimental operations. Then, we theoretically discussed the most rational approach to incorporating PI and empirically validated this theory. The introduction of PI enhanced model performance across different datasets: on the synthetic datasets, the average reduction in $E_1$ was about $0.84 \times 10^{-2}$, whereas on the experimental datasets, $E_1$ decreased slightly, with an average reduction of approximately $0.06 \times 10^{-2}$. Subsequently, to address issues at the edges of the results, we further employed the Softplus activation function, which improved the continuity at the edge of the reconstruction profile because its characteristics aligning with the properties of the physical quantities in question. It was also observed that the Softplus activation function improved model performance: on the synthetic datasets, the average reduction in $E_1$ was about $1.06 \times 10^{-2}$ on average, while on the experimental datasets, $E_1$ decreased by about $0.11 \times 10^{-2}$ on average. Finally, we compared and analyzed the impact of the PILF. The adoption of PILF effectively constrained the predictions, ensuring they better satisfied the inherent experimental constraints and resulted in better outcomes. For the Exp_EAST dataset with $c_1$ set to 0.618, after applying PILF, the average decrease in $E_2$ is $0.95 \times 10^{-2}$. For the Exp_HL-2A dataset with $c_1$ set to 1.0, the application of PILF led to a more substantial average reduction in $E_2$, amounting to $4.59 \times 10^{-2}$.

The work demonstrates the enhancement effect of the physics-informed deep learning model architecture on models, indicating its potential to improve the accuracy of surrogate models in future applications related to rapid profile reconstruction for between-shot analysis in nuclear fusion diagnostics, which are crucial for real-time control and decision-making processes in fusion reactors. However, the model has its limitations. The training data lacks uncertainty information from experimental measurements, preventing the model from engaging in uncertainty-related research and analysis. Additionally, the model's PI and PILF only incorporate line integration information and do not encompass MHD-related knowledge, limiting the model's capabilities in MHD-unstable discharge scenarios.

Incorporating updated backbone networks, an expanded dataset, and higher-quality data will likely lead to further enhancements in model performance. Including experimental data with uncertainty information will bolster the rigor and completeness of model result analysis and discussion. More work should be done to prove the applicability of the model for different phases of a discharge, such as limiter and diverted scenarios, peaked and hollow emissivity distributions, etc. We can expand our model's capabilities by integrating a broader spectrum of physics-informed knowledge, including MHD insights and even information from other diagnostics. By employing a mixture of experts (MOE) architecture and federated learning strategies, we can significantly enhance the model's performance in handling different experimental discharge scenarios. These will be conducted in the future study. The experiment data during the current study are not publicly available for legal/ethical reasons but are available from the corresponding author on reasonable request.

## Acknowledgments

The authors express their sincere gratitude to Dr. Liqing Xu and Dr. Chaowei Mai from the Hefei Institute of Plasma Physics, Chinese Academy of Sciences, for their invaluable assistance and support throughout this research. Special thanks are also extended to Dr. Dong Li from the Southwest Institute of Physics, China National Nuclear Corporation, and to Dr. Xianli Huang and Cong Zhang from the ENN Science and Technology Development Co., Ltd. for their contributions and encouragement. Their expertise and guidance have been instrumental to the success of this work. This work is supported by the National Natural Science Foundation of China (No. 12405266) and National Natural Science Foundation of China (No.52406198). The code, models, and some datasets will be publicly available at [https://github.com/calledice/onion](https://github.com/calledice/onion).

## References

1. Seo, J. *et al.* Development of an operation trajectory design algorithm for control of multiple 0D parameters using deep reinforcement learning in KSTAR. *Nucl. Fusion* **62**, 086049 (2022).

2. Seo, J. *et al.* Multimodal Prediction of Tearing Instabilities in a Tokamak. in *2023 International Joint Conference on Neural Networks (IJCNN)* 1–8 (IEEE, Gold Coast, Australia, 2023). doi:10.1109/IJCNN54540.2023.10191359.

3. Vega, J. *et al.* Disruption prediction with artificial intelligence techniques in tokamak plasmas. *Nat. Phys.* **18**, 741–750 (2022).

4. Fu, Y. *et al.* Machine learning control for disruption and tearing mode avoidance. *Physics of Plasmas* **27**, 022501 (2020).

5. Liu, Z. Y. *et al.* Prediction of fishbone linear instability in tokamaks with machine learning methods. *Nucl. Fusion* **65**, 016007 (2025).

6. Bustos, A., Ascasíbar, E., Cappa, A. & Mayo-García, R. Automatic identification of MHD modes in magnetic fluctuation spectrograms using deep learning techniques. *Plasma*

*Phys. Control. Fusion* **63**, 095001 (2021).

7. Kaptanoglu, A. A. *et al.* Exploring data-driven models for spatiotemporally local classification of Alfvén eigenmodes. *Nucl. Fusion* **62**, 106014 (2022).

8. Wei, Y., Levesque, J. P., Hansen, C., Mauel, M. E. & Navratil, G. A. MHD mode tracking using high-speed cameras and deep learning. *Plasma Phys. Control. Fusion* **65**, 074002 (2023).

9. Lee, J. E., Seo, P. H., Bak, J. G. & Yun, G. S. A machine learning approach to identify the universality of solitary perturbations accompanying boundary bursts in magnetized toroidal plasmas. *Sci Rep* **11**, 3662 (2021).

10. Han, W. *et al.* Estimating cross-field particle transport at the outer midplane of TCV by tracking filaments with machine learning. *Nucl. Fusion* **63**, 076025 (2023).

11. Yang, K. N. *et al.* Neural network identification of the weakly coherent mode in I-mode discharge on EAST. *Nucl. Fusion* **64**, 016035 (2024).

12. Piccione, A., Berkery, J. W., Sabbagh, S. A. & Andreopoulos, Y. Physics-guided machine learning approaches to predict the ideal stability properties of fusion plasmas. *Nucl. Fusion* **60**, 046033 (2020).

13. Liu, Y., Lao, L., Li, L. & Turnbull, A. D. Neural network based prediction of no-wall $\beta_N$ limits due to ideal external kink instabilities. *Plasma Phys. Control. Fusion* **62**, 045001 (2020).

14. G. Dong *et al.* Deep learning based surrogate models for first-principles global simulations of fusion plasmas. *Nucl. Fusion* **61**, 126061 (2021).

15. Li, H., Fu, Y., Li, J. & Wang, Z. Machine learning of turbulent transport in fusion

plasmas with neural network. *Plasma Sci. Technol.* **23**, 115102 (2021).

16. Van Mulders, S. *et al.* Rapid optimization of stationary tokamak plasmas in RAPTOR: demonstration for the ITER hybrid scenario with neural network surrogate transport model QLKNN. *Nucl. Fusion* **61**, 086019 (2021).

17. Clement, M. D., Logan, N. C. & Boyer, M. D. Neoclassical toroidal viscosity torque prediction via deep learning. *Nucl. Fusion* **62**, 026022 (2022).

18. Kates-Harbeck, J., Svyatkovskiy, A. & Tang, W. Predicting disruptive instabilities in controlled fusion plasmas through deep learning. *Nature* **568**, 526–531 (2019).

19. Raissi, M., Perdikaris, P. & Karniadakis, G. E. Physics-informed neural networks: A deep learning framework for solving forward and inverse problems involving nonlinear partial differential equations. *Journal of Computational Physics* **378**, 686–707 (2019).

20. Jalili, D. *et al.* Physics-informed neural networks for heat transfer prediction in two-phase flows. *International Journal of Heat and Mass Transfer* **221**, 125089 (2024).

21. Liu, M., Liang, L. & Sun, W. A generic physics-informed neural network-based constitutive model for soft biological tissues. *Computer Methods in Applied Mechanics and Engineering* **372**, 113402 (2020).

22. Wessels, H., Weißenfels, C. & Wriggers, P. The neural particle method – An updated Lagrangian physics informed neural network for computational fluid dynamics. *Computer Methods in Applied Mechanics and Engineering* **368**, 113127 (2020).

23. Sahli Costabal, F., Yang, Y., Perdikaris, P., Hurtado, D. E. & Kuhl, E. Physics-Informed Neural Networks for Cardiac Activation Mapping. *Front. Phys.* **8**, 42 (2020).

24. Ameya D. Jagtap, A. D. J. & George Em Karniadakis, G. E. K. Extended Physics-

Informed Neural Networks (XPINNs): A Generalized Space-Time Domain Decomposition Based Deep Learning Framework for Nonlinear Partial Differential Equations. *CICP* **28**, 2002–2041 (2020).

25. Wang, Z. *et al.* Deep Learning Based Surrogate Model a fast Soft X-ray (SXR) Tomography on HL-2 a Tokamak. *J Fusion Energ* **43**, 52 (2024).

26. Mai, C. *et al.* Application of deep learning to soft x-ray tomography at EAST. *Plasma Phys. Control. Fusion* **64**, 115009 (2022).

27. Ferreira, D. R. Applications of Deep Learning to Nuclear Fusion Research. Preprint at http://arxiv.org/abs/1811.00333 (2018).

28. Nagayama, Y. Tomography of $m$ =1 mode structure in tokamak plasma using least-square-fitting method and Fourier–Bessel expansions. *Journal of Applied Physics* **62**, 2702–2706 (1987).

29. Chen, K. *et al.* 2-D soft x-ray arrays in the EAST. *Review of Scientific Instruments* **87**, 063504 (2016).

30. Blatzheim, M., Böckenhoff, D., & the Wendelstein 7-X Team. Neural network regression approaches to reconstruct properties of magnetic configuration from Wendelstein 7-X modeled heat load patterns. *Nucl. Fusion* **59**, 126029 (2019).

31. Matos, F., Svensson, J., Pavone, A., Odstrčil, T. & Jenko, F. Deep learning for Gaussian process soft x-ray tomography model selection in the ASDEX Upgrade tokamak. *Review of Scientific Instruments* **91**, 103501 (2020).

32. Simonyan, K. & Zisserman, A. Very Deep Convolutional Networks for Large-Scale Image Recognition. Preprint at https://doi.org/10.48550/arXiv.1409.1556 (2015).

33. Tianbo Wang. Reconstruction of soft X-ray and tungsten concentration profiles in Tokamaks using Bayesian method. (Ghent University, 2019).

34. He, K., Zhang, X., Ren, S. & Sun, J. Deep Residual Learning for Image Recognition. Preprint at http://arxiv.org/abs/1512.03385 (2015).

35. Vaswani, A. *et al.* Attention Is All You Need. Preprint at http://arxiv.org/abs/1706.03762 (2023).

36. Dosovitskiy, A. *et al.* An Image is Worth 16x16 Words: Transformers for Image Recognition at Scale. Preprint at http://arxiv.org/abs/2010.11929 (2021).

37. Li, D. *et al.* Bayesian tomography and integrated data analysis in fusion diagnostics. *Rev. Sci. Instrum.* **87**, 11E319 (2016).

38. Kirkpatrick, S., Gelatt, C. D. & Vecchi, M. P. Optimization by Simulated Annealing. *Science* **220**, 671–680 (1983).

39. Loshchilov, I. & Hutter, F. SGDR: Stochastic Gradient Descent with Warm Restarts. Preprint at https://doi.org/10.48550/arXiv.1608.03983 (2017).

40. Smith, L. N. & Topin, N. Super-Convergence: Very Fast Training of Neural Networks Using Large Learning Rates. Preprint at https://doi.org/10.48550/arXiv.1708.07120 (2018).

41. Wu, T. *et al.* Compositional Generative Inverse Design. Preprint at http://arxiv.org/abs/2401.13171 (2024).

42. Liu, Z. *et al.* Plasma electron density profile tomography for EAST based on integrated data analysis. *Nucl. Fusion* **64**, 126006 (2024).

43. Kingma, D. P. & Ba, J. Adam: A Method for Stochastic Optimization. Preprint at

https://doi.org/10.48550/arXiv.1412.6980 (2017).

# Appendix

## A. Adam Optimizer and Cosine Annealing Learning Rate Scheduler

In this work, we employ a combination of the Adam optimizer and a cosine annealing learning rate scheduler to optimize the model during training. Below is a detailed explanation of the mechanism behind these two components.

**Adam Optimizer**

The Adam optimizer (Adaptive Moment Estimation) is an adaptive learning rate optimization algorithm that computes adaptive learning rates for each parameter[43]. It combines the advantages of two other extensions of stochastic gradient descent: Momentum and RMSprop.

Adam works by maintaining two moving averages for each parameter: First moment ($\hat{m}_t$): This is the exponentially decaying average of past gradients (i.e., the momentum term). Second moment ($\hat{v}_t$): This is the exponentially decaying average of past squared gradients (i.e., the variance of the gradients).

At each iteration, Adam adjusts the learning rate based on the magnitude of the gradients, making it suitable for sparse gradients or noisy updates. The parameter update rule in Adam is given by:

$$\theta_t = \theta_{t-1} - \frac{\eta}{\sqrt{\hat{v}_t} + \epsilon} \hat{m}_t$$

Where:

- $\theta_t$ is the parameter at time step $t$.
- $\eta$ is the global learning rate (initially set to 0.0001 in our case).
- $\hat{m}_t$ is the bias-corrected first moment estimate.
- $\hat{v}_t$ is the bias-corrected second moment estimate.
- $\epsilon$ is a small constant (e.g., $10^{-8}$ ) to prevent division by zero.

The optimizer's self-adaptive mechanism adjusts the learning rate for each parameter based on its individual gradients, which helps to stabilize the training process and speeds up convergence, especially in the presence of sparse or fluctuating gradients.

**Cosine Annealing Learning Rate Scheduler**

The cosine annealing learning rate scheduler dynamically adjusts the global learning rate throughout training, following a cosine curve. The learning rate decreases from a maximum value to a minimum value over the course of a defined number of epochs (or training steps), without abrupt changes, which is known to improve convergence in deep learning models.

The learning rate $\eta_t$ at each time step $t$ is given by:

$$\eta_t = \eta_{\min} + \frac{1}{2}(\eta_{\max} - \eta_{\min})\left(1 + \cos\left(\frac{T_{\text{cur}}}{T_{\max}}\pi\right)\right)$$

Where:

- $\eta_{\max}$ is the initial learning rate (set to 0.0001).
- $\eta_{\min}$ is the minimum learning rate (set to 0.00001).
- $T_{\mathrm{cur}}$ is the current epoch.
- $T_{\max}$ is the maximum number of epochs, in this case, 50.

The scheduler starts with the maximum learning rate and gradually reduces it following a cosine curve over the course of 50 epochs. This strategy allows the optimizer to take larger steps in the early stages of training, enabling rapid learning, and smaller steps towards the end to fine-tune the model and prevent overshooting the optimal solution.

**Interaction Between Adam and the Scheduler**

The Adam optimizer adaptively adjusts the learning rate for each parameter based on $\hat{m}_t$ and $\hat{v}_t$. However, these adjustments are scaled by the globally modified learning rate $\eta_t$, which undergoes cosine decay as dictated by the scheduler. The parameter update rule in Adam with cosine annealing learning rate scheduler is given by:

$$\theta_t = \theta_{t-1} - \frac{\eta_t}{\sqrt{\hat{v}_t} + \epsilon}\hat{m}_t$$

The combination of Adam and the cosine annealing scheduler results in an effective optimization strategy that benefits from both components' strengths: Adam: Adam adjusts the learning rate on a per-parameter basis, dynamically scaling the step size according to the gradient history, ensuring stable updates and improving convergence. Cosine Annealing Scheduler: The scheduler modulates the global learning rate over time, ensuring that the learning rate starts high and progressively decays to a smaller value, which allows the model to explore the parameter space in the beginning and fine-tune the parameters in the later stages of training.

Thus, the Adam optimizer provides fine-grained control over each parameter's learning rate, while the cosine annealing scheduler smoothly decays the global learning rate to encourage convergence without abrupt jumps. The changes in the global learning rate during model training are shown in the **Figure 22**.

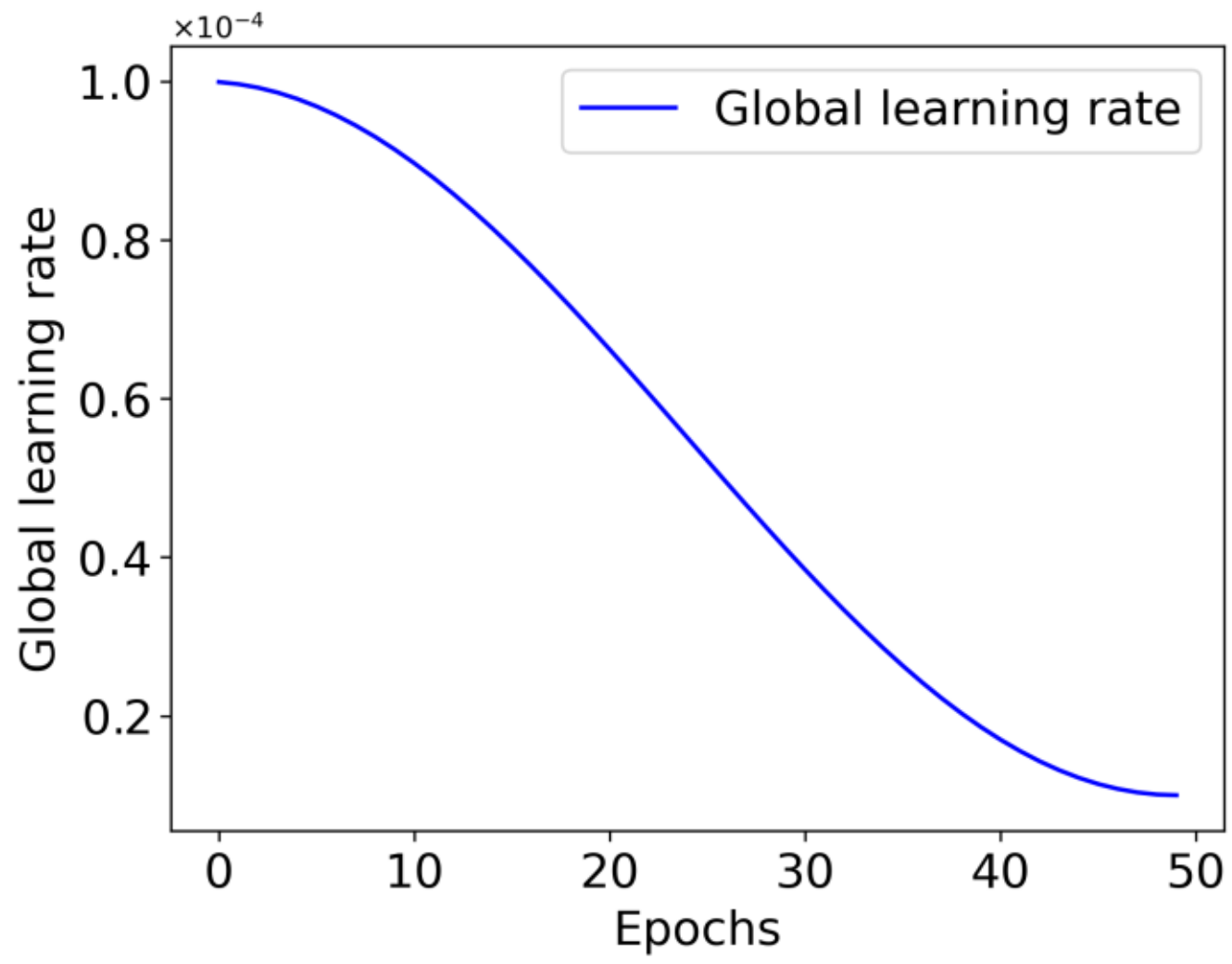

**Figure 22** Global learning rate during model training

Training Configuration

- Optimizer: Adam, with an initial learning rate of 0.0001.
- Cosine Annealing Scheduler: Period of 50 epochs, with a minimum learning rate of 0.00001.

- Training Duration: 50 epochs.

By combining the benefits of Adam's adaptive learning rates with the smooth decay provided by the cosine annealing scheduler, this training strategy enables efficient model optimization, ensuring both fast convergence and long-term stability throughout the training process. **Figure 23** illustrates the loss curves for model training on EXP_EAST using the Adam optimizer alone versus using the Adam optimizer in conjunction with a cosine annealing learning rate scheduler. The loss curve of the former exhibits oscillation and may not have converged by 50 epochs, whereas the latter shows a smoother loss curve that has essentially reached convergence by the 50-epoch mark.

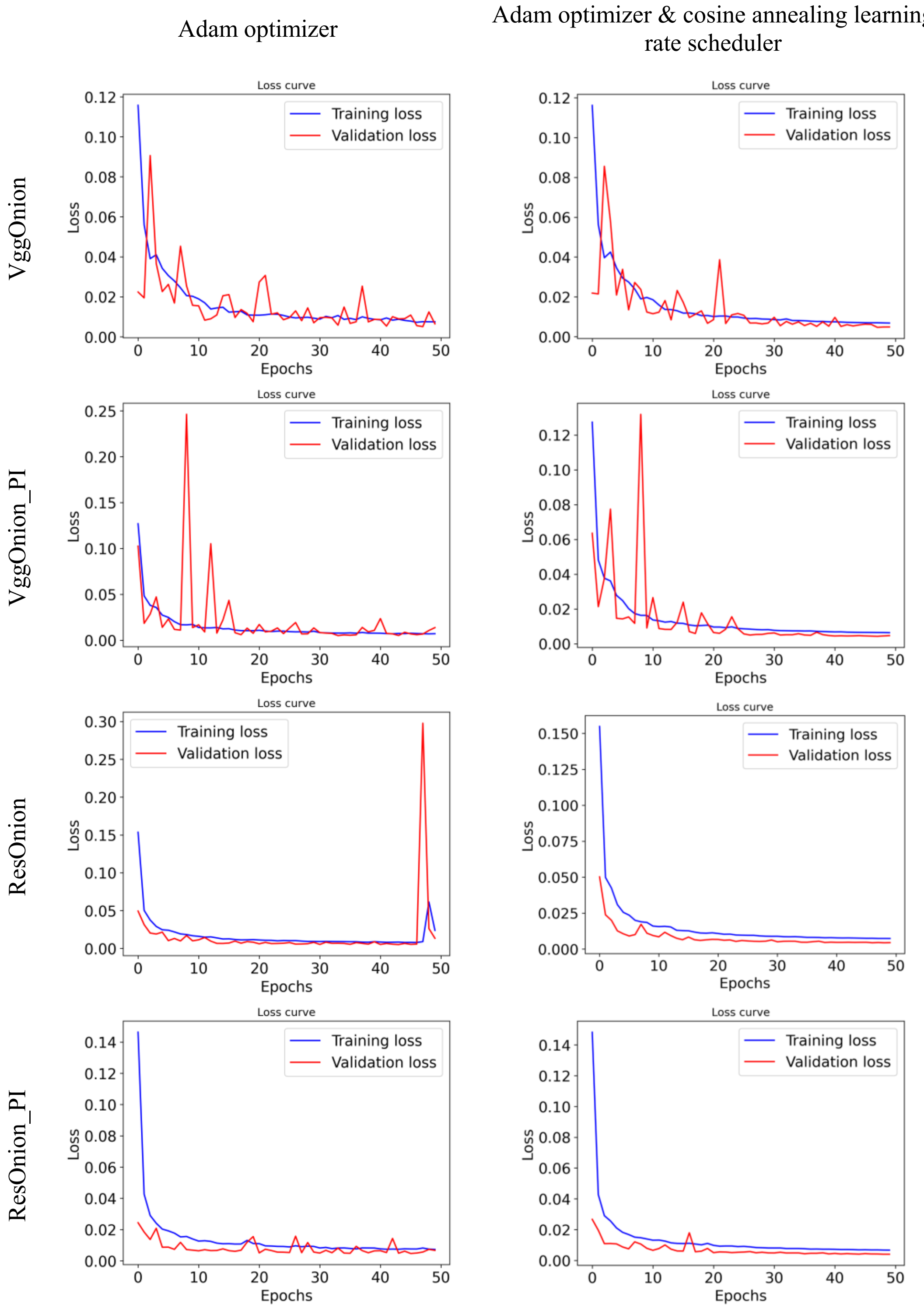

**Figure 23** Loss curves for model training on EXP_EAST.

## B. Loss curves for cases in Section 4

This appendix presents the loss curves for the cases discussed in **Section 4**, illustrating that after training for 50 epochs, the model described in this paper has essentially reached a state of convergence. The loss curves in the following figures provide visual evidence of the model's performance over the course of training, showing how the loss decreases with each epoch. By the 50th epoch, the trend indicates that the model's learning process has stabilized, suggesting that further training may not improve the model's performance. This observation supports the conclusion that the proposed model achieves satisfactory convergence within the specified number of epochs.

**Cases in Section 4.1:**

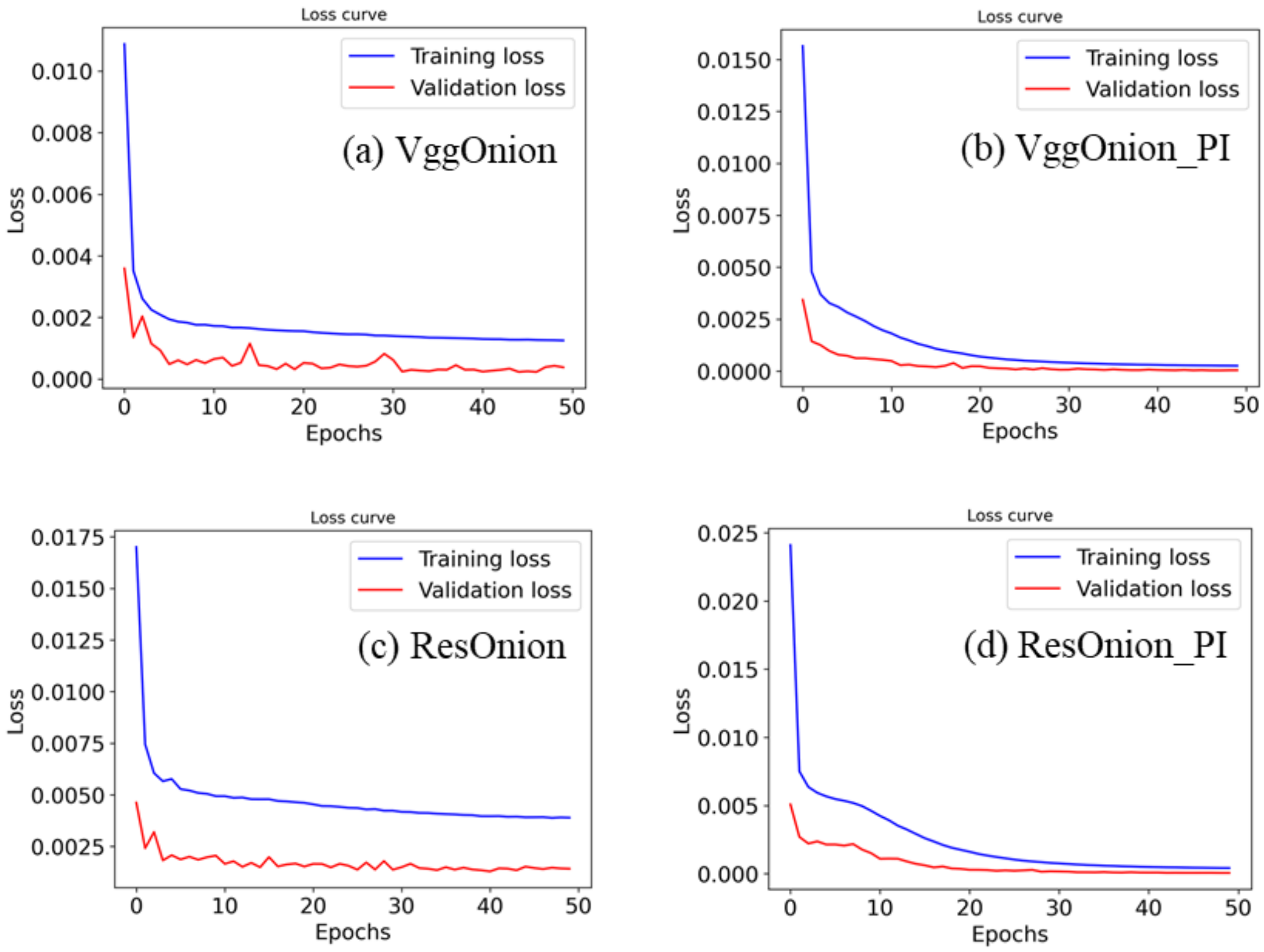

**Figure 24** Loss curves for four models training on Synthetic_EAST.

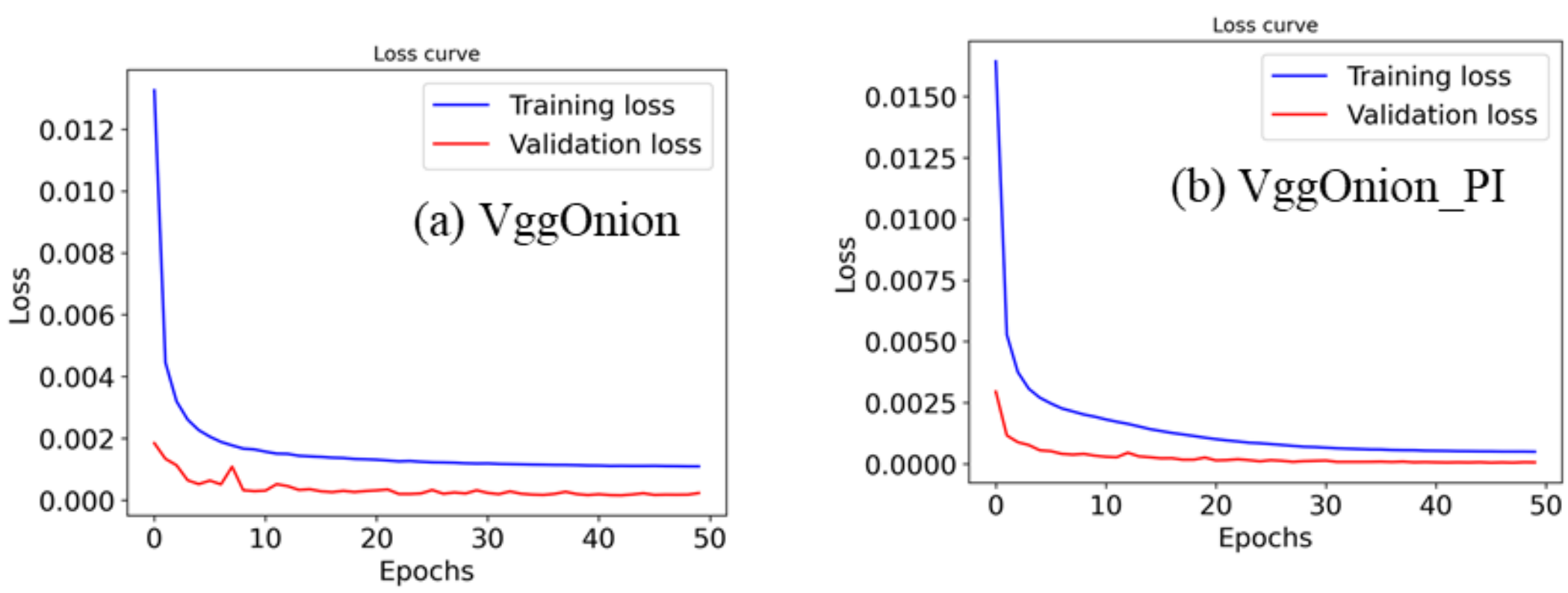

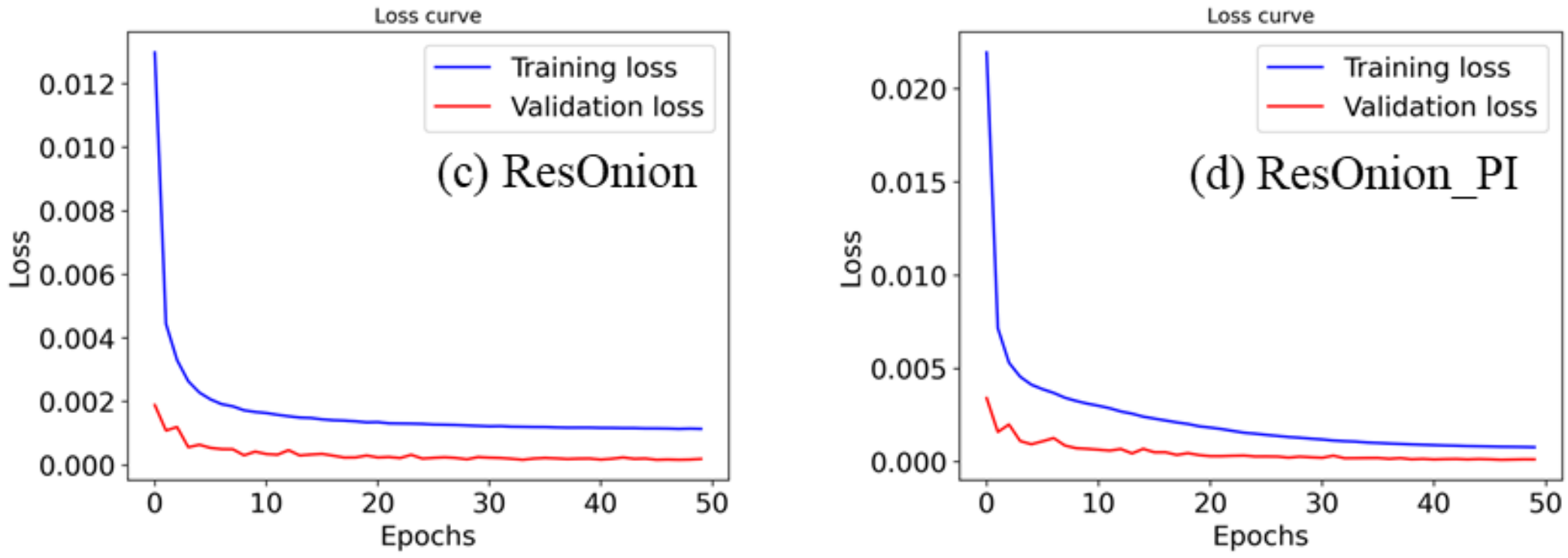

**Figure 25** Loss curves for four models training on Synthetic_HL-2A.

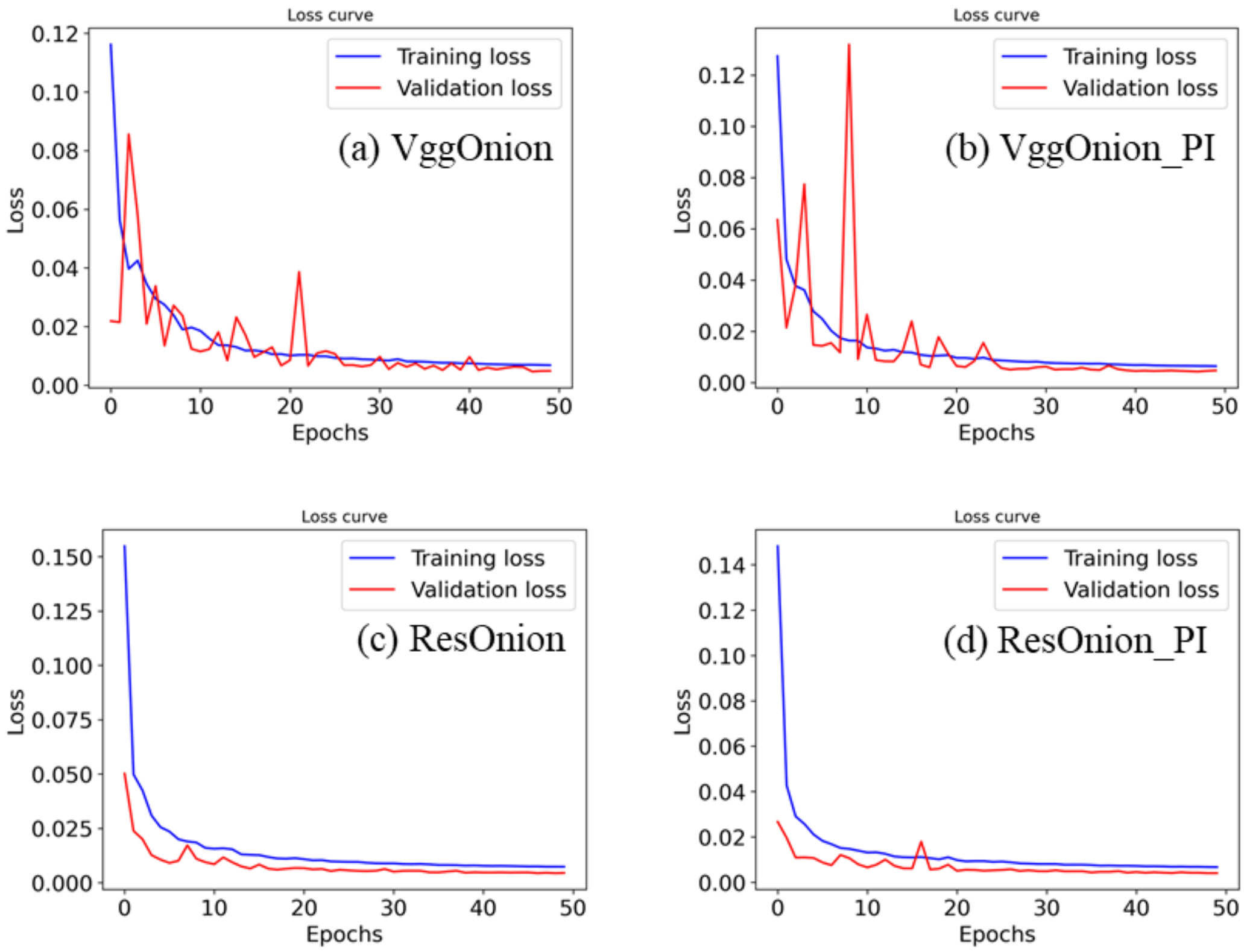

**Figure 26** Loss curves for four models training on Exp_EAST.

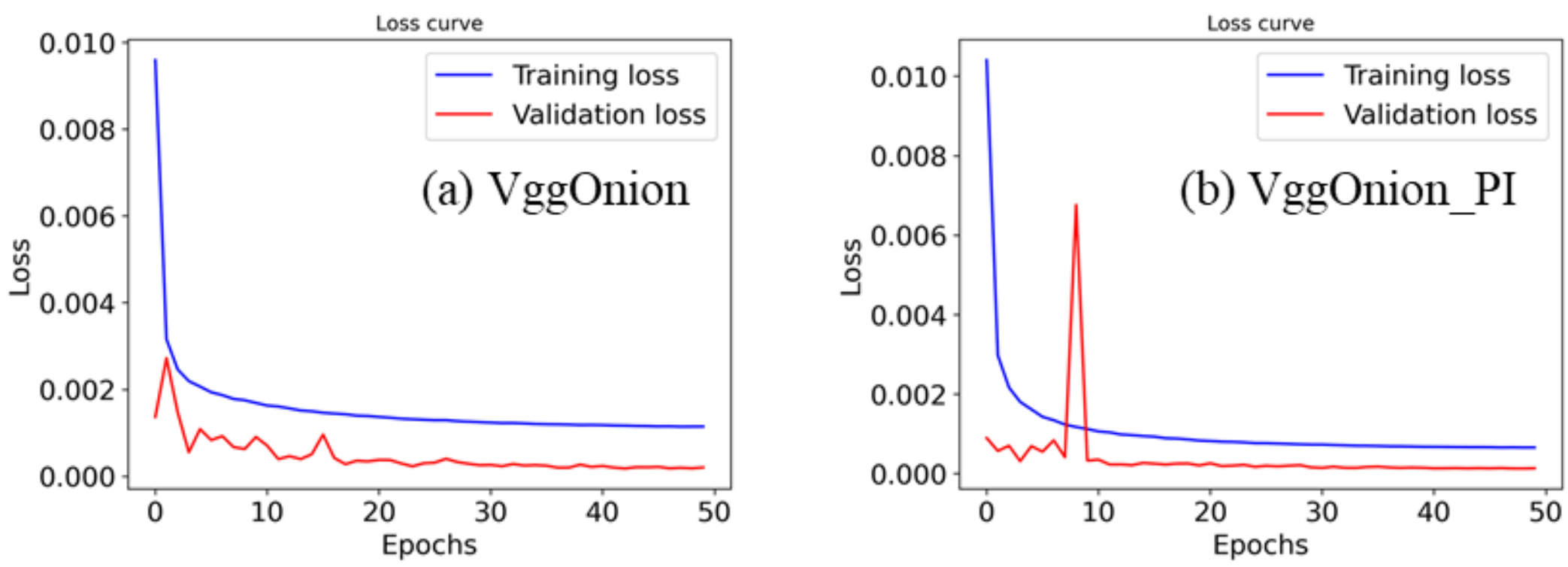

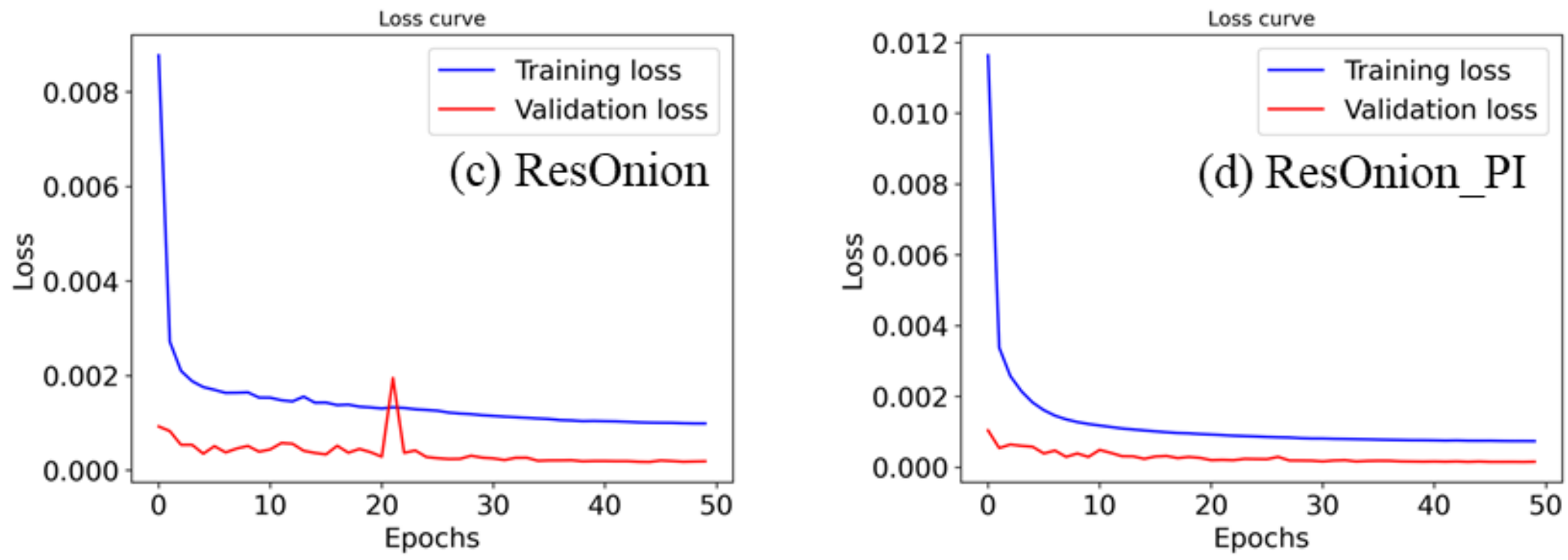

**Figure 27** Loss curves for four models training on Exp_HL-2A.

**Cases in Section 4.2:**

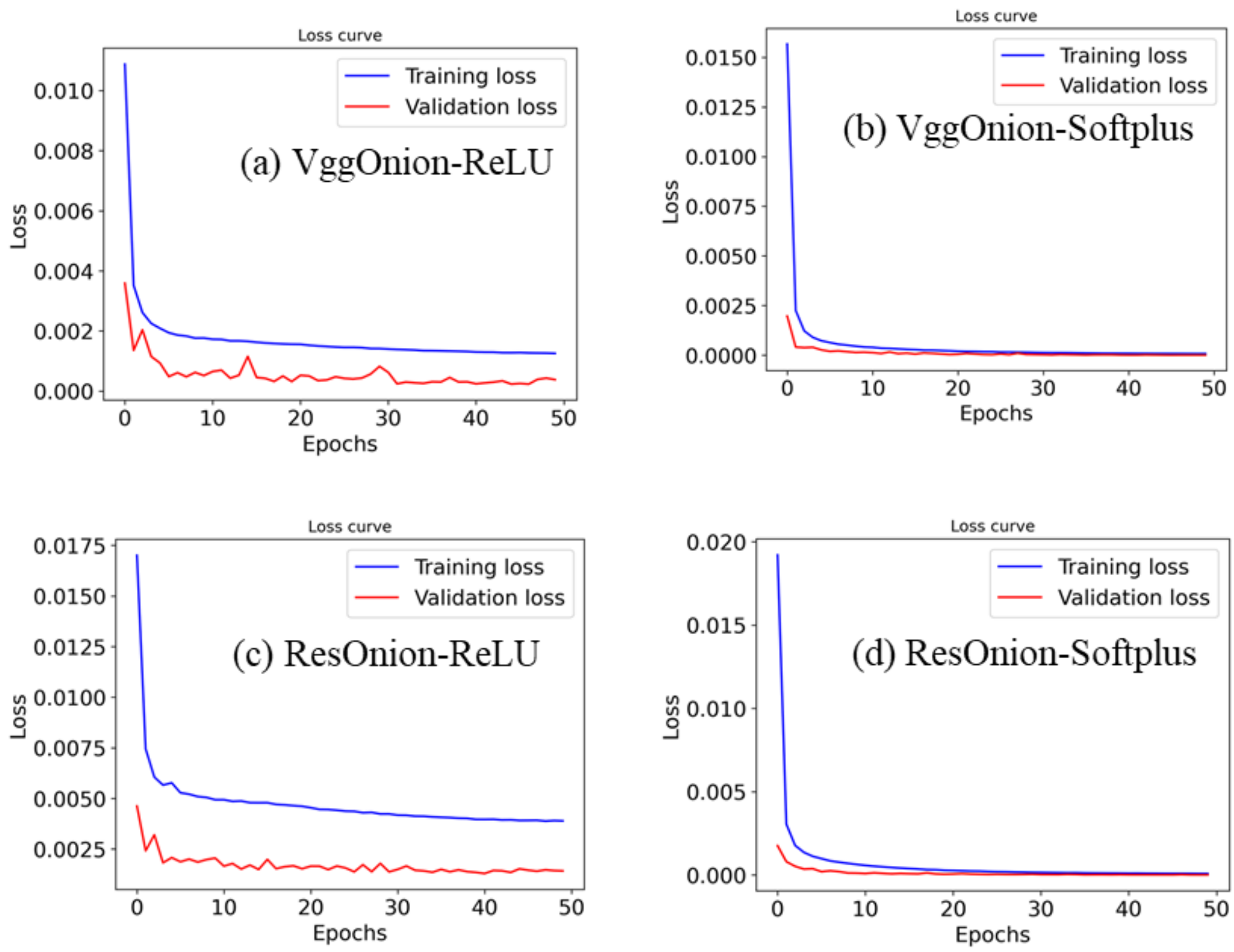

**Figure 28** Loss curves for four models training on Synthetic_EAST.

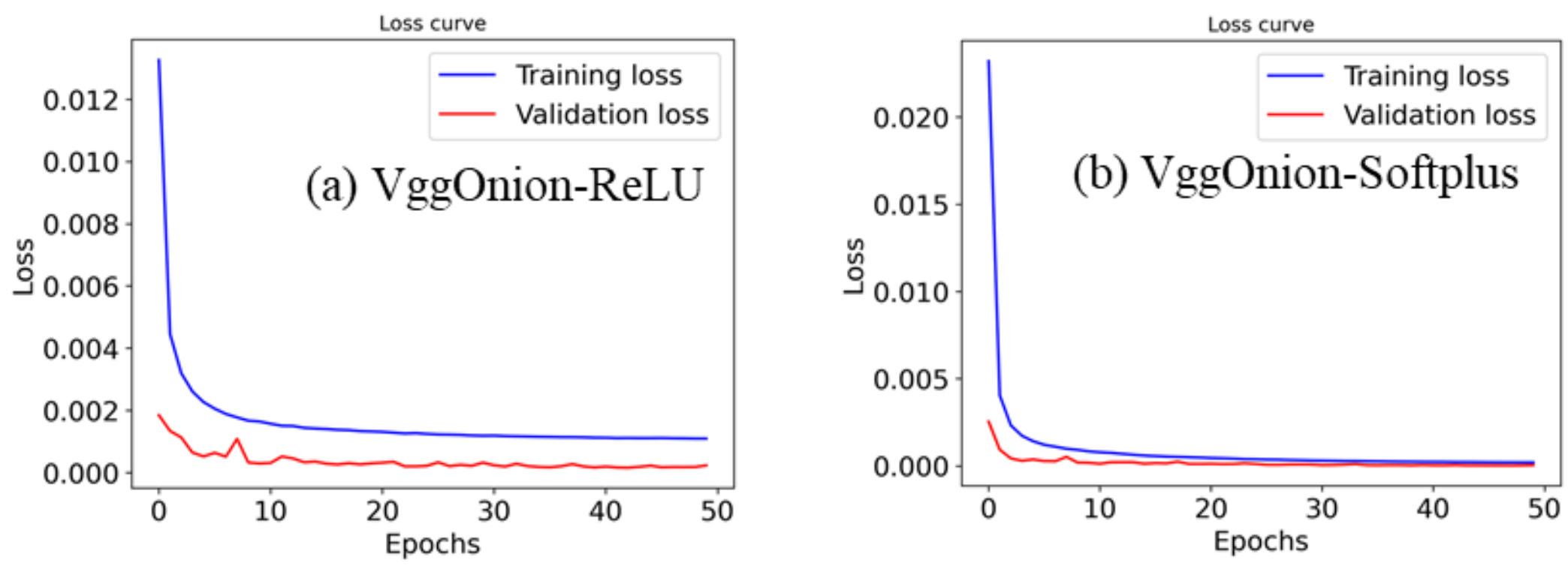

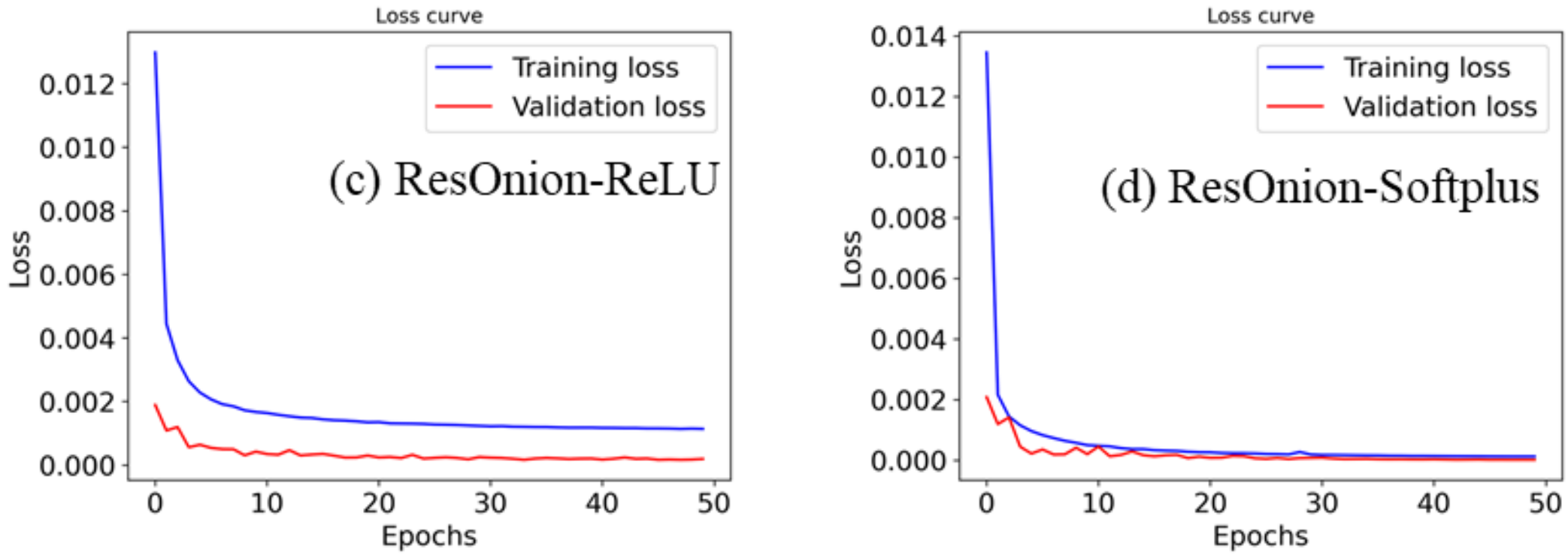

**Figure 29** Loss curves for four models training on Synthetic_HL-2A.

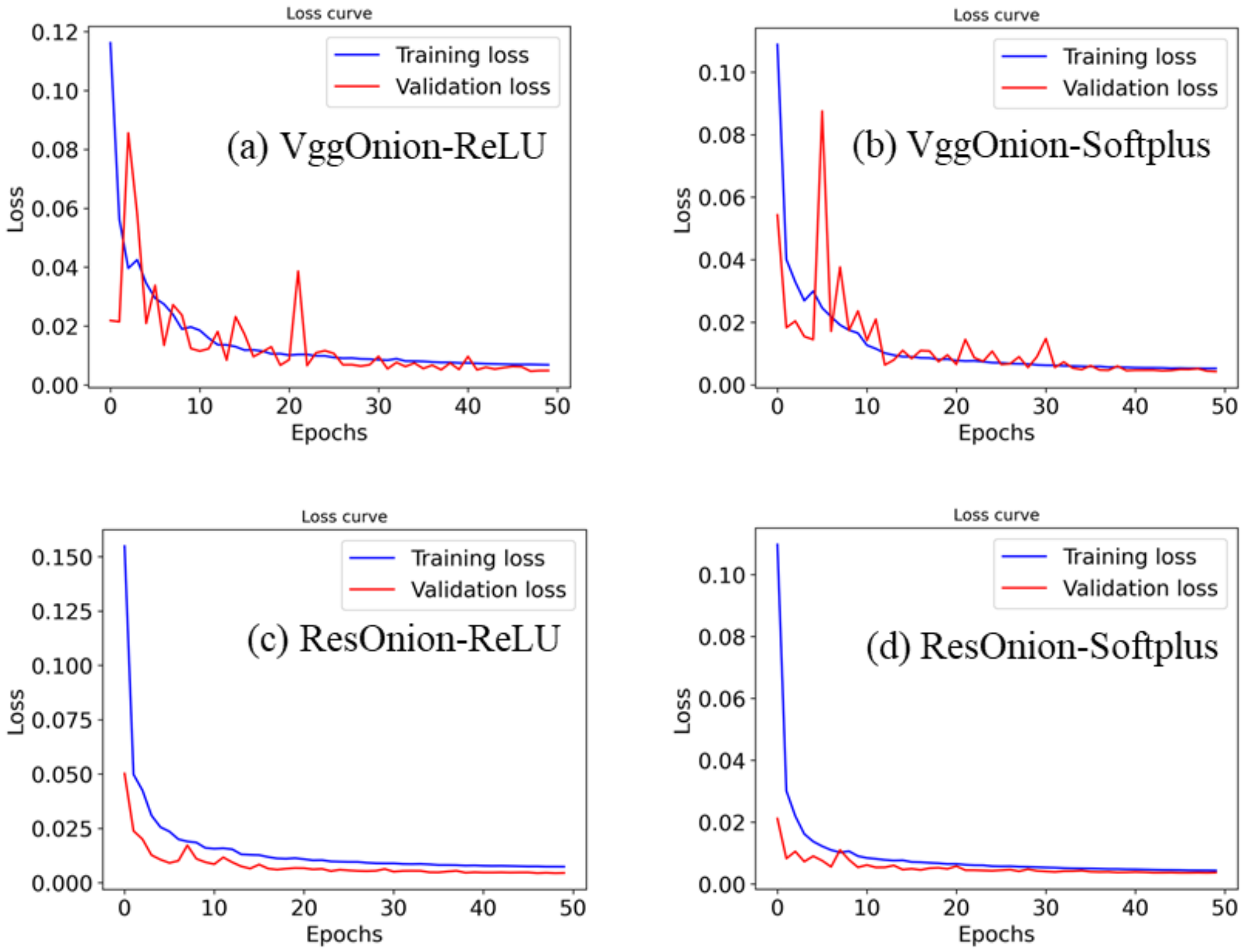

**Figure 30** Loss curves for four models training on Exp_EAST.

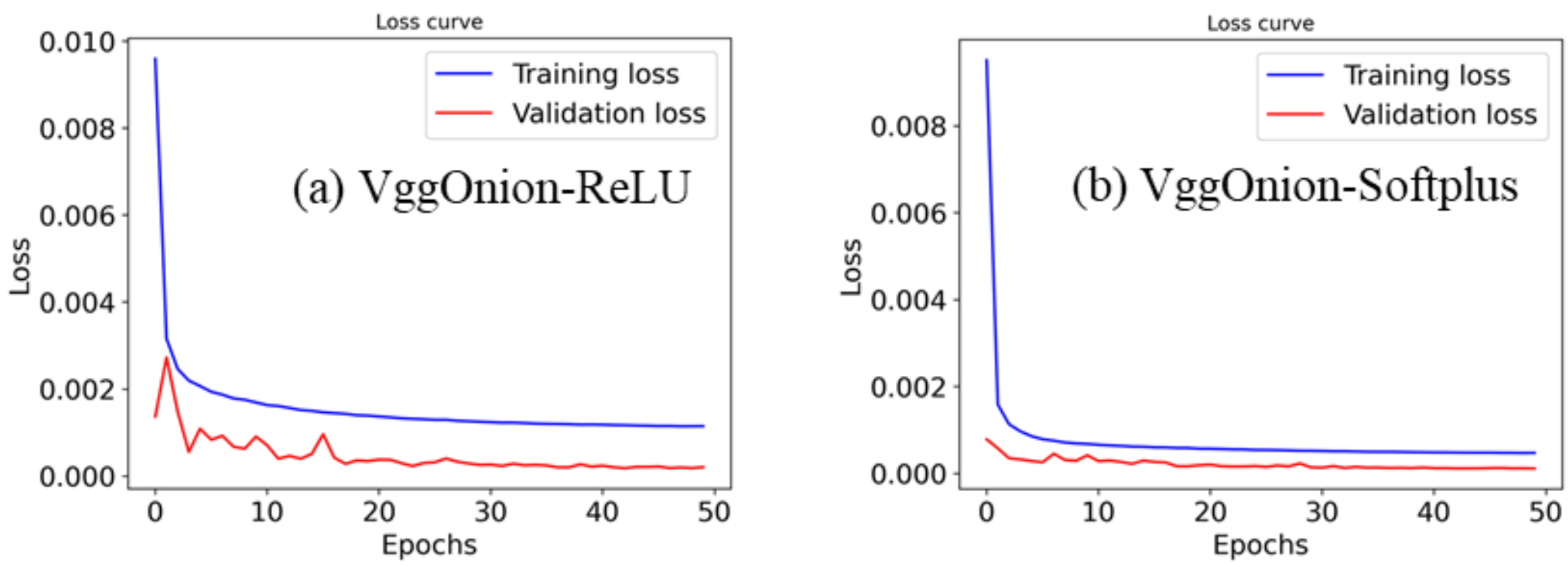

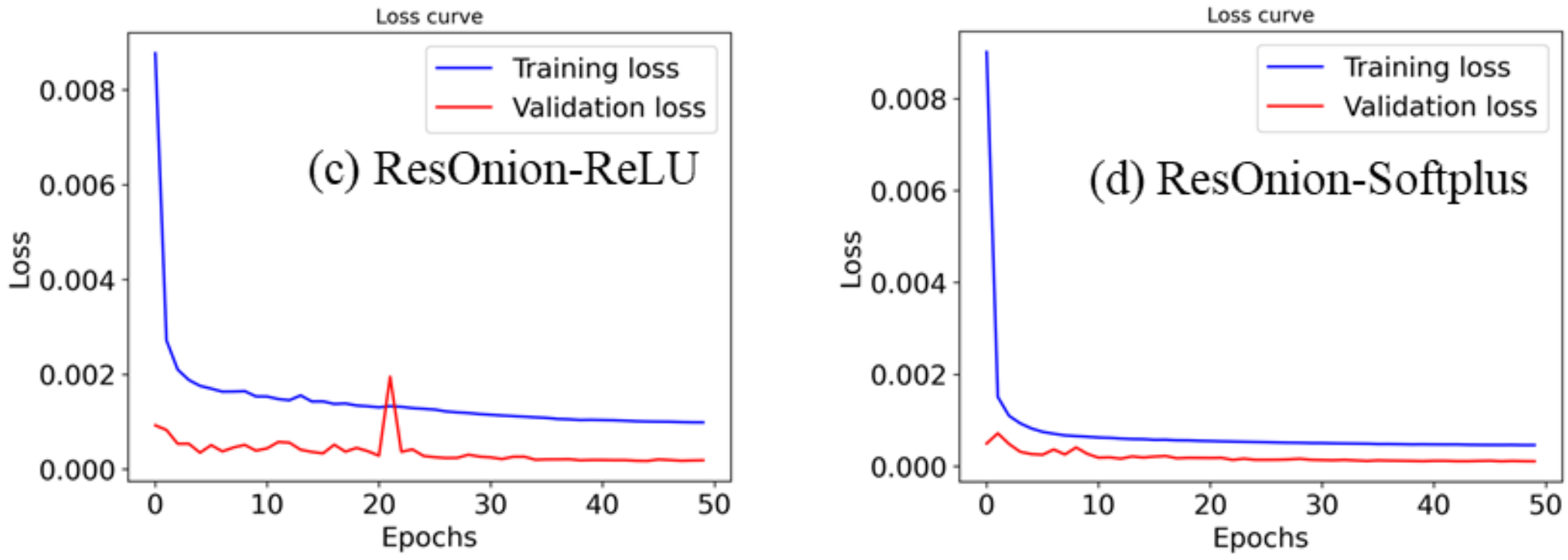

**Figure 31** Loss curves for four models training on Exp_HL-2A.

**Cases in Section 4.3:**

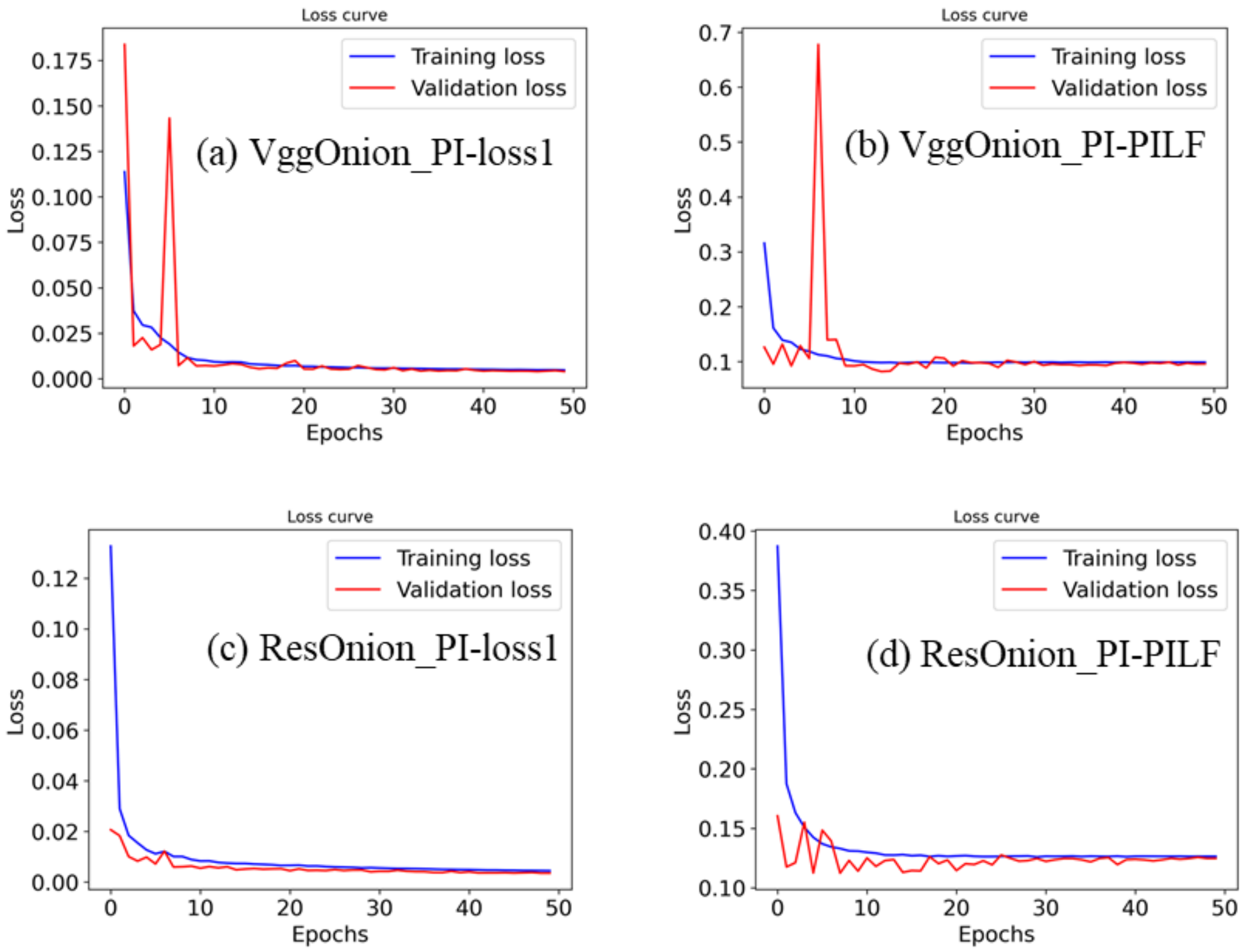

**Figure 32** Loss curves for four models training on Exp_EAST.

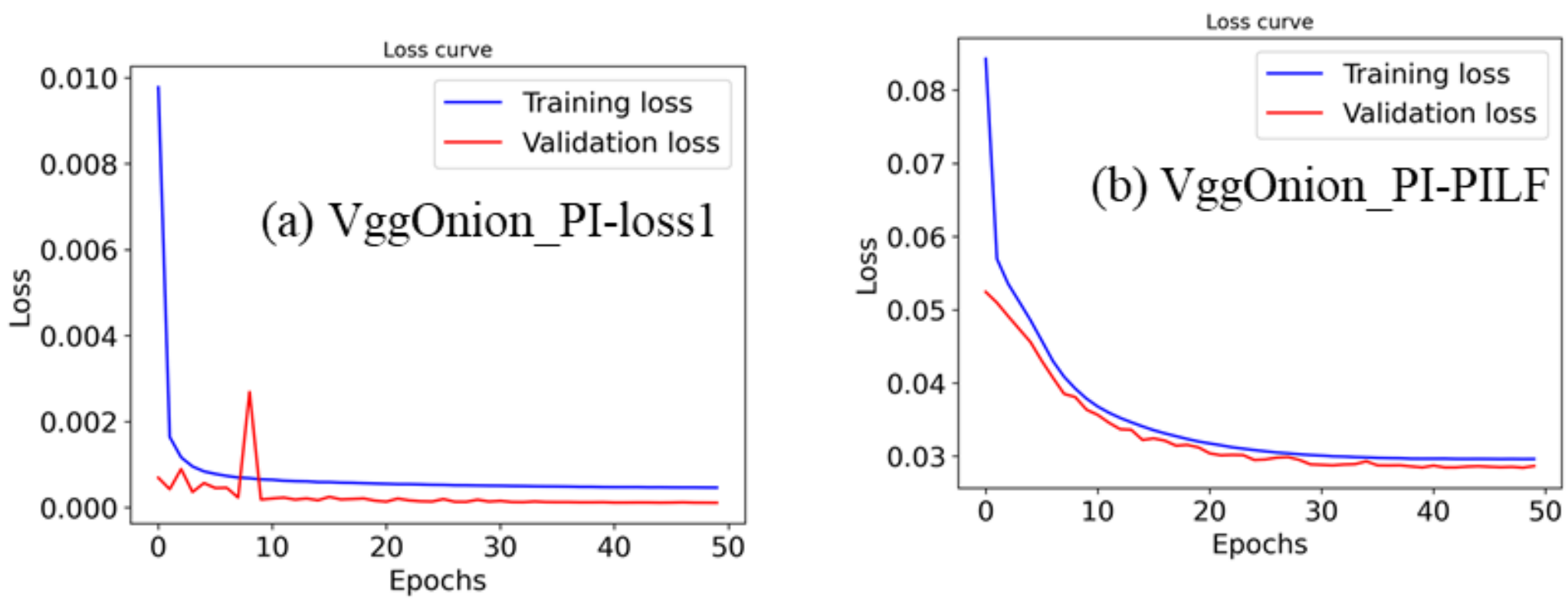

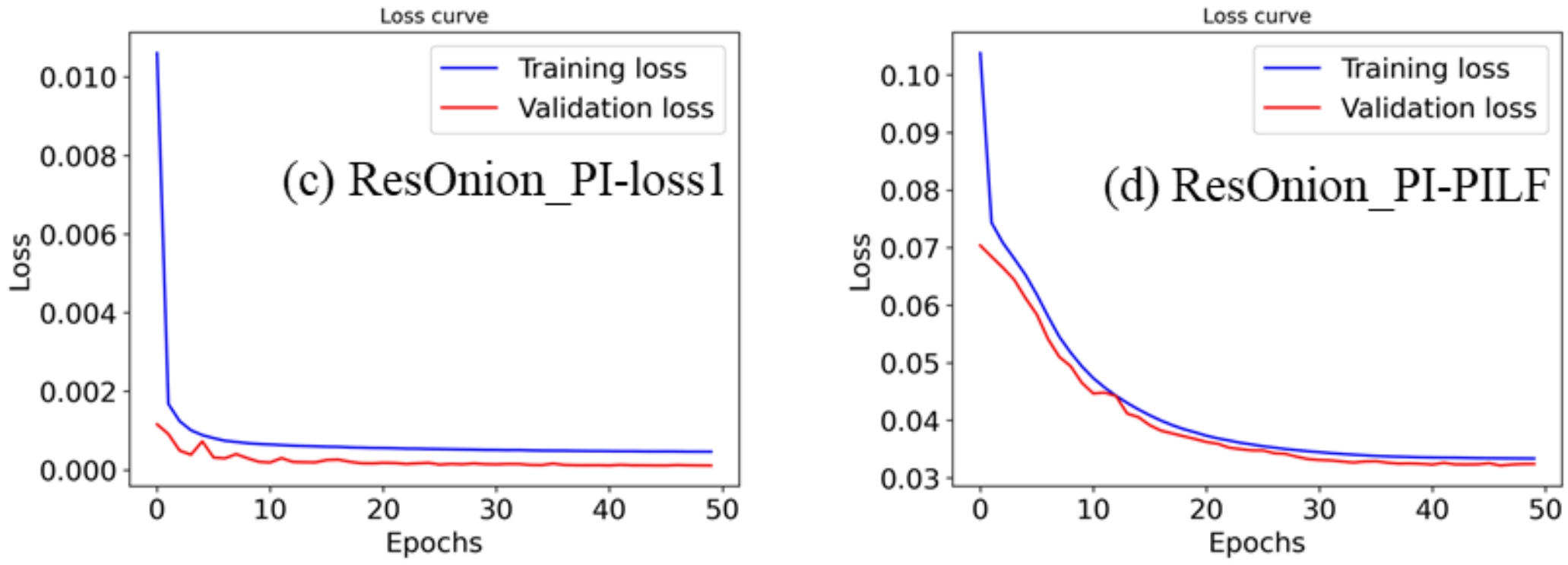

**Figure 33** Loss curves for four models training on Exp_HL-2A.